\documentclass{article} %
\usepackage[preprint]{colm2026_conference}

\usepackage{microtype}
\usepackage{graphicx}
\usepackage{booktabs} %

\usepackage{lineno}
\usepackage{hyperref}

\usepackage{amsmath}
\usepackage{amssymb}
\usepackage{mathtools}
\usepackage{amsthm}
\usepackage{bbm}
\usepackage{comment}
\usepackage{caption}
\usepackage{subcaption}
\usepackage{pifont}
\usepackage{tabularx}
\usepackage{colortbl}
\usepackage{url}
\usepackage{algorithm}
\usepackage{algorithmic}
\usepackage{enumitem}
\usepackage{wrapfig}

\usepackage[skins,breakable]{tcolorbox}
\usepackage[normalem]{ulem}

\usepackage[capitalize,noabbrev]{cleveref}

\definecolor{darkblue}{rgb}{0, 0, 0.5}
\hypersetup{colorlinks=true, citecolor=darkblue, linkcolor=darkblue, urlcolor=darkblue}

\theoremstyle{plain}

\theoremstyle{definition}

\theoremstyle{remark}

\usepackage[textsize=tiny]{todonotes}

\newcommand{\qrhead}{\textsc{QRHead}}
\newcommand{\rethead}{retrieval head}
\newcommand{\methodname}{\textsc{DySCO}}

\newcommand{\pathtrans}{\textsc{Path Traversal}}
\newcommand{\mrcr}{\textsc{MRCR}}
\newcommand{\longbench}{\textsc{LongBenchV2}}
\newcommand{\clipper}{\textsc{Clipper}}

\newcommand{\hotpot}{\textsc{Hotpot}}
\newcommand{\infqa}{\textsc{InfQA}}
\newcommand{\infmc}{\textsc{InfMC}}

\newcommand{\attnts}{\textsc{UniAttnS}}

\definecolor{light-purple}{RGB}{151,156,171}
\definecolor{blue-color}{RGB}{40,166,189}
\definecolor{pink-color}{RGB}{237,46,104} 
\definecolor{dark-grey-color}{RGB}{79,91,102}

\newcommand{\param}[1]{
\textcolor{pink-color}{\small{\texttt{\detokenize{#1}}}}
}

\newcommand{\prompttext}[1]{\textcolor{black}{#1}}

\newtcolorbox[list inside=prompt,auto counter,number within=section]{prompt}[1][]{
    colbacktitle=black!80,
    colframe=black!80,
    coltitle=white,
    fontupper=\footnotesize,
    boxsep=5pt,
    left=0pt,
    right=0pt,
    top=0pt,
    bottom=0pt,
    boxrule=1pt,
    enhanced, 
    breakable,
    skin first=enhanced,
    skin middle=enhanced,
    skin last=enhanced,
    #1,
}

\definecolor{exsinputcolor}{HTML}{E4F2DA}
\definecolor{exsoutputcolor}{HTML}{EEE2FB}
\newtcolorbox[list inside=prompt,auto counter,number within=section]{exsinput}[1][]{
    colback=exsinputcolor,
    colbacktitle=black!80,
    colframe=black!80,
    coltitle=white,
    fontupper=\footnotesize,
    boxsep=5pt,
    left=0pt,
    right=0pt,
    top=0pt,
    bottom=0pt,
    boxrule=1pt,
    enhanced, 
    breakable,
    skin first=enhanced,
    skin middle=enhanced,
    skin last=enhanced,
    #1,
}

\newtcolorbox{exsoutput}[1][]{
    colback=exsoutputcolor,
    colbacktitle=black!80,
    colframe=black!80,
    coltitle=white,
    fontupper=\footnotesize,
    boxsep=5pt,
    left=0pt,
    right=0pt,
    top=0pt,
    bottom=0pt,
    boxrule=1pt,
    enhanced, 
    breakable,
    skin first=enhanced,
    skin middle=enhanced,
    skin last=enhanced,
    #1,
}

\title{\textsc{DySCO}: Dynamic Attention-Scaling Decoding for \\ Long-Context Language Models}

\author{Xi Ye$^{\spadesuit}$\thanks{Equal contribution.} \quad Wuwei Zhang$^{\spadesuit*}$ \quad Fangcong Yin$^{\diamondsuit}$ \quad Howard Yen$^{\spadesuit}$ \quad Danqi Chen$^{\spadesuit}$ \\
$^{\spadesuit}$ Princeton Language and Intelligence, Princeton University \\
$^{\diamondsuit}$ Department of Computer Science, New York University \\
\texttt{\{xi.ye, wuwei.zhang\}@princeton.edu}
}

\begin{document}

\ifcolmsubmission
\linenumbers
\fi

\maketitle

\begin{abstract}

Understanding and reasoning over long contexts is a crucial capability for language models (LMs).
Although recent models support increasingly long context windows, their accuracy often deteriorates as input length grows.
In practice, models often struggle to keep attention aligned with the most relevant context throughout decoding.
In this work, we propose \methodname{}, a novel decoding
algorithm for improving long-context reasoning. \methodname{} leverages \emph{retrieval heads}—a
subset of attention heads specialized for long-context
retrieval—to identify task-relevant tokens at each
decoding step and explicitly up-weight them. By doing so,
\methodname{} dynamically adjusts attention during
generation to better utilize relevant context.
The method is training-free and can be applied directly to any off-the-shelf LMs.
Across multiple instruction-tuned and reasoning models,
\methodname{} consistently improves performance on
challenging long-context reasoning benchmarks, yielding
relative gains of up to 25\% on \mrcr{} and \longbench{} at
128K context length with modest additional compute.
Further analysis highlights the importance of both dynamic
attention rescaling and retrieval-head guided
selection for the effectiveness of the method, while
providing interpretability insights into decoding-time
attention behavior.
Our code is available at \url{https://github.com/princeton-pli/DySCO}. 
\end{abstract}

\section{Introduction}

\begin{wrapfigure}{r}{0.45\textwidth}
\vspace{-3em}
    \centering
    \includegraphics[width=\linewidth,trim=0 720 1100 0,clip]{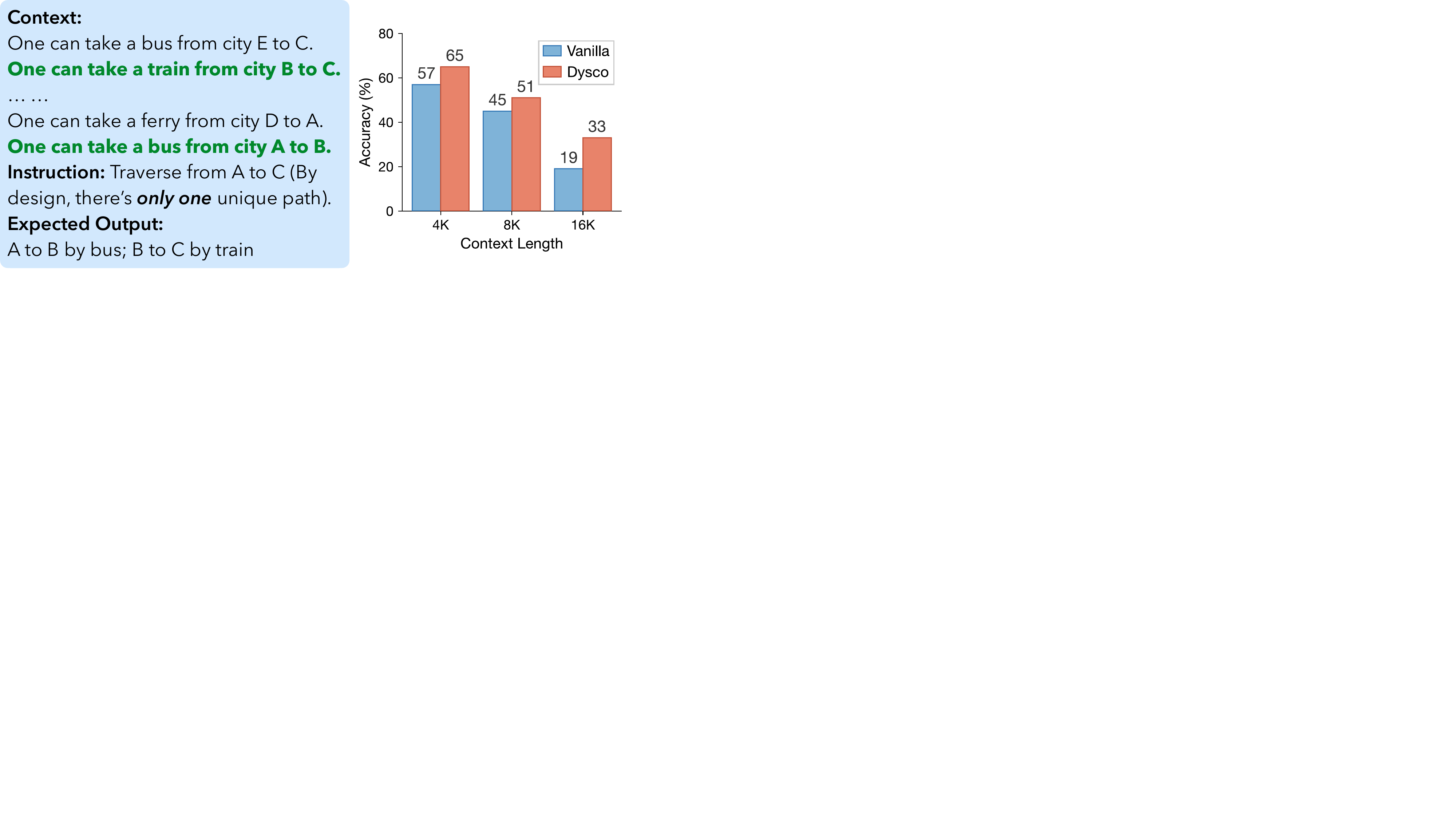}
    \caption{Top: An illustrative \emph{Path Traversal} task (simplified). Solving the task requires dynamically locating relevant context during decoding. Bottom: Accuracy as a function of context length for Qwen3-8B with and without \methodname{}. Despite the total context being only 16K tokens, both models exhibit severe performance degradation as context length increases.}
    \label{fig:intro_challenge}
\end{wrapfigure}

Recent advances in language models (LMs) have enabled
processing of extremely long context windows, unlocking
applications such as repository-level code understanding
and long-document question answering. Driven by improvements in data curation and transformer architectures,
modern LMs now support context lengths of 128K tokens
and beyond~\citep{Comanici2025Gemini2P,Yang2025Qwen3TR,Bai2025KimiKO,gpt52,claude_46}.
However, model performance often degrades
significantly as input length increases, even on simple
tasks, commonly known as ``context rot''~\citep{hong2025context,reallyhardlongcontext}.
Figure~\ref{fig:intro_challenge} illustrates this phenomenon using a simple \pathtrans{} task~\citep{Longproc}, which requires LMs to trace a path in a graph where each node has exactly one outgoing edge. Recent models such as Qwen3-8B and Llama-3.1-8B see an accuracy drop from around
 60\% at 4K tokens to below 20\% at 16K tokens, despite supporting context lengths of up to 128K tokens.

\begin{figure*}[t]
    \centering
    \includegraphics[width=1.0\linewidth,trim=0 170 0 100,clip]{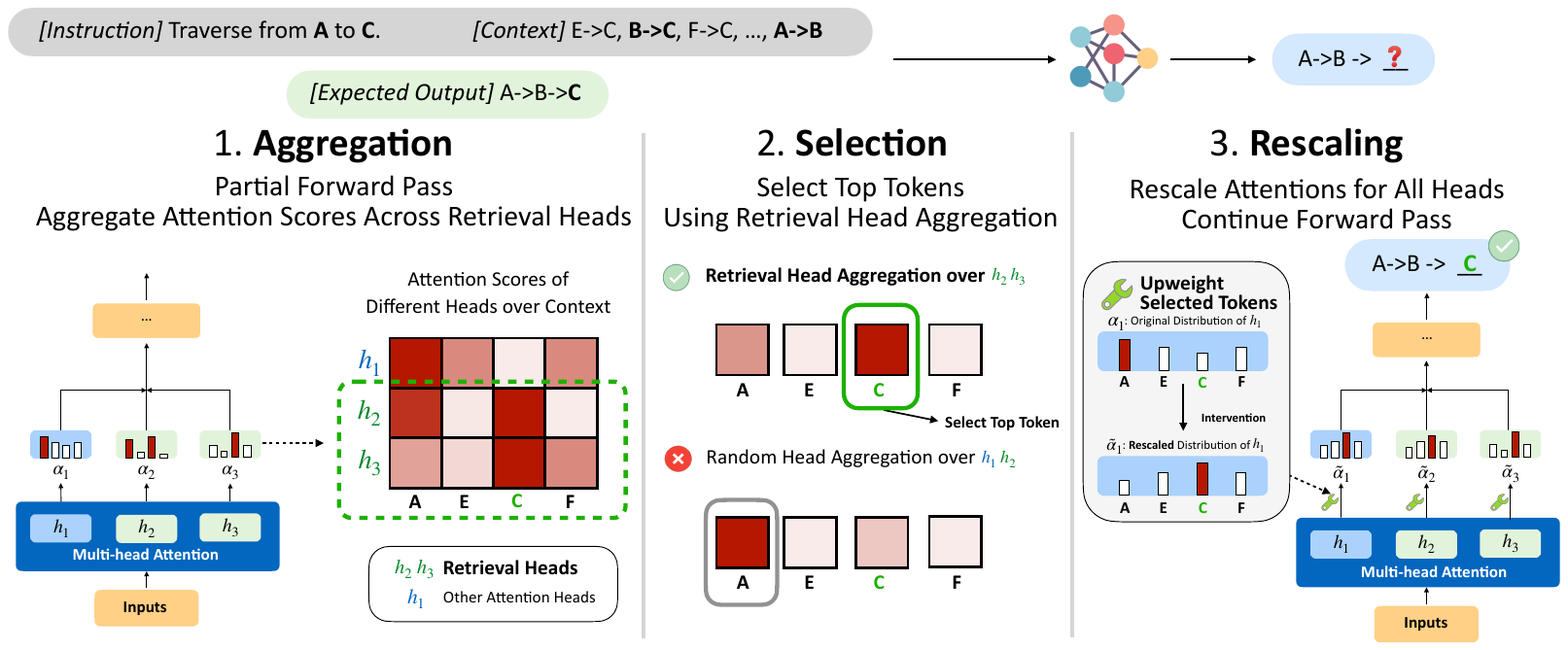}
    \caption{Overview of \methodname{} algorithm. At each decoding step, \methodname{} consists of three stages: (1) \textbf{Aggregation}: We run a partial forward pass over the input sequence to obtain attentions of retrieval heads, such as \qrhead{}, and use them to assign relevance scores to context tokens; (2) \textbf{Selection}: We use the relevance scores to select the important tokens; (3) \textbf{Rescaling}: We up-weight the important tokens by intervening attention logits of all attention heads and run a full forward pass to sample the next token.}
    \label{fig:algo_overview}
\end{figure*}

We attribute this degradation to a key challenge in how LMs
utilize long and information-dense contexts during generation: effective long-context reasoning requires models to
continuously focus attention on the most task-relevant parts
of the context as the generation state evolves. 
In practice, however, vanilla decoding often fails to exhibit this behavior over
long contexts.
For example, in the \pathtrans{} task shown in
Figure~\ref{fig:intro_challenge}, LMs must iteratively
identify the next edge from the context based on the
current node. Our analysis (\S\ref{sec:qrhead_analysis})
shows that attention is often insufficiently focused on
the relevant context at each generation step, leading to
errors even when the required information is present.
In this work, we propose a novel decoding
algorithm, \methodname{} (\textbf{Dy}namic Attention-\textbf{S}caling De\textbf{CO}ding), which dynamically adjusts attention on the fly during generation. \methodname{} is lightweight, training-free, and directly applicable to off-the-shelf language models. \methodname{} operates exclusively at the decoding stage, after the long-context prefilling phase that accounts for the majority of computation. As a result, it introduces only small additional overhead—for example, approximately 4\% extra FLOPs when generating 8K tokens from 128K-token inputs.

The key idea of \methodname{} is to identify relevant tokens at each decoding step using \emph{retrieval heads}~\citep{wu2025retrieval,qrheads},
and then upweight attention to these tokens during generation.
Retrieval heads are a specialized subset (1-2\%) of attention heads that are responsible for long-context retrieval: compared to other
heads, they assign higher and more stable attention to
context that is critical for next-token prediction.
Notably, we find that retrieval heads remain consistently
focused on relevant tokens even when overall model
performance degrades with increasing context length
(\S\ref{sec:qrhead_analysis}).
After prefilling, \methodname{} operates in three stages at each decoding
step (Figure~\ref{fig:algo_overview}). 1) \emph{aggregation}: we run a partial forward pass (\S\ref{sec:algo}) to aggregate attention scores from
retrieval heads; 2) \emph{selection}: we
identify a small set of context tokens with the highest
aggregated attention scores; 3) \emph{rescaling}:
we upweight attention to the selected tokens by
intervening on the attention logits of \emph{all} heads
and run a full forward pass to generate the next token.

\methodname{} introduces a lightweight mechanism for \textit{dynamic} attention shaping during decoding. Prior work on attention scaling largely relies on \emph{static} patterns. For example, long-context extension methods uniformly rescale all attention logits by a constant factor~\citep{peng2024yarn,chen2026critical,Nakanishi2025ScalableSoftmaxIS}. Other approaches impose fixed, position-dependent patterns~\citep{hsiehfound,zhang2024found} to mitigate the ``lost-in-the-middle'' issue~\citep{liu-etal-2024-lost}, which is less pronounced in modern LMs (e.g., Qwen3-8B performs strongly on RULER~\citep{Yang2025Qwen3TR}).
In contrast, our method performs \emph{dynamic, token-selective} scaling driven by internal signals from retrieval heads. Retrieval heads have also been previously used for improving long-context training pipeline through data curation~\citep{qiu-etal-2025-eliciting}, auxiliary objectives~\citep{Liu2025MuDAFLM}, or localized interventions~\citep{Zhu2025FocusDM}.
Our work shows that they can be seamlessly integrated into the decoding procedure, leading to strong improvements in long-context reasoning without any additional training.

We evaluate \methodname{} on both instruction-tuned and reasoning LMs across a wide range of long-context tasks. \methodname{} consistently improves performance on challenging long-context reasoning benchmarks. Notably, for Qwen3-8B, \methodname{} delivers up to 25\% relative improvements on \mrcr{}~\citep{mrcr} and \longbench{}~\citep{longbenchv2} compared to YaRN alone, with $\le$4\% additional FLOPs. Further analysis highlights the importance of both dynamic scaling and the use of retrieval heads for identifying important tokens.
In summary, we introduce dynamic attention scaling, a new way of shaping attention in long-context LMs.
Our results demonstrate that \methodname{} leads to consistent gains in long-context reasoning while offering mechanistic insights into LMs' long-context behavior.

\section{Background and Motivation}

We first set up the preliminaries of retrieval heads~\citep{wu2025retrieval,qrheads}, an important building block of our approach (\S\ref{sec:intro_qrhead}). We then study \pathtrans{}, a synthetic task designed to stress-test basic long-context reasoning, and use it to reveal the connection between retrieval heads and long-context reasoning capabilities (\S\ref{sec:qrhead_analysis}).

\subsection{Preliminaries: Retrieval Heads}
\label{sec:intro_qrhead}

We consider an autoregressive transformer LM $\mathcal{M}$~\citep{Vaswani+2017} that generates the next token $x_{t+1}$ conditioned on the prefix $x_{\le t}$, which includes both the input context and previously generated tokens. Let $\mathcal{H}=\{h_i\}$ denote the set of attention heads in $\mathcal{M}$. At decoding step $t$, each head $h \in \mathcal{H}$ produces an attention distribution: $\boldsymbol{\alpha}_t^{(h)} \in \mathbb{R}^{t}$,  where $\alpha_{t,i}^{(h)}$ is the attention mass for token $x_i \in \boldsymbol{x}_{\le t}$.

\paragraph{Retrieval Heads.}
\citet{wu2025retrieval} discovered a universal set of attention heads that exhibit copy-like behavior during decoding, which they term \emph{retrieval heads}. Concretely, when generating token $x_t$, a retrieval head $h$ concentrates its attention on a prior occurrence of the same token in the context. That is, $\alpha^{(h)}_{t,i}$ is high for some $i < t$ such that $x_i = x_t$. These heads are typically sparse (less than 5\% of all heads) and provide a mechanistic explanation for how language models perform explicit token lookup and copy-paste from long context.

\paragraph{Query-Focused Retrieval Heads (\qrhead{}).}

Retrieval heads can be identified using different criteria~\citep{qrheads,qiu-etal-2025-eliciting,Zhu2025FocusDM}. 
Our method builds on \qrhead{}~\citep{qrheads}, which identifies attention heads based on query–context attention mass using examples from realistic long-context tasks. We adopt \qrhead{} due to its strong generalization across both diverse domains in the BEIR re-ranking benchmark~\citep{thakur2021beir} and a range of challenging long-context reasoning tasks. Additional details on \qrhead{} are provided in Appendix~\ref{app:detail_qrhead}.

\subsection{Retrieval Heads Stay Focused on Relevant Context}
\label{sec:qrhead_analysis}

\begin{wrapfigure}{rt}{0.49\textwidth}
\vspace{-2em}
    \centering
    \includegraphics[width=1.0\linewidth,trim=0 0 0 0,clip]{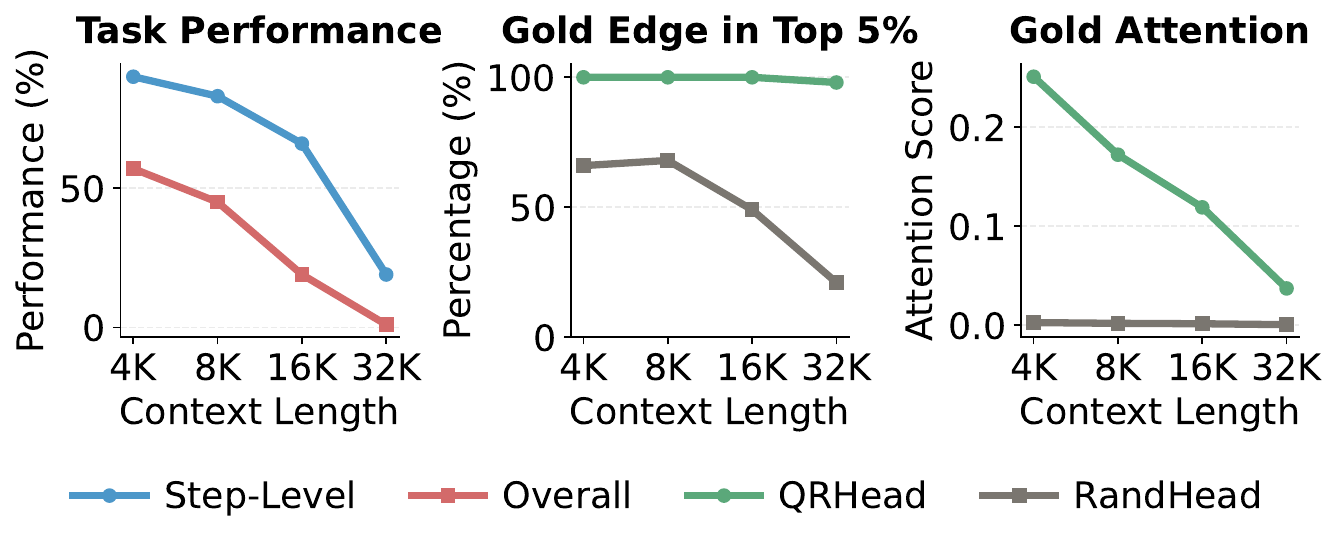}
    \vspace{-1em}
    \caption{\textbf{Left}: Performance of Qwen3-8B on \pathtrans{} as context length increases.
\textbf{Middle}: Fractions that the gold edge appear among the top-5\% edges ranked by attention score (sum of attention over all tokens in the span) from \qrhead{} versus random heads.
\textbf{Right}: Attention mass assigned to gold edges by \qrhead{} and random heads.
Despite severe performance degradation and a reduction in attention mass on gold edges, \qrhead{} consistently allocates substantially higher attention to the gold edges.
}
    \label{fig:motivation}
\end{wrapfigure}

\paragraph{Diagnostic task: \pathtrans{}.} 
Long-context reasoning requires dynamically attending to relevant information at each decoding step, conditioned not only on the inputs but also on the intermediate state encoded in $x_{\le t}$.  To analyze this capability in a controlled setting, we use a synthetic task, \pathtrans{}~\citep{Longproc}. \pathtrans{} takes a list of edges and requires finding a path from a start to a target node. The graph is constructed so that each node on the gold path has exactly one outgoing edge, reducing the task to iteratively retrieving the correct next node (a deterministic process explicitly specified in the prompt). We control context length (up to 32K) by varying the number of nodes. Despite its simplicity, performance degrades sharply with longer context: step-level accuracy drops from near-perfect to $\sim$20\% at 32K tokens (Figure~\ref{fig:motivation}).
This highlights the need for repeated, dynamic retrieval as a core challenge of long-context reasoning. Further details are provided in Appendix~\ref{app:detail_qr_analysis}.

\paragraph{Behavior of retrieval heads.}
We further analyze the behavior of retrieval heads, which provides insight into LMs' failure modes on long-context reasoning tasks. We partition the context into edge-level spans and compute the sum of attention mass assigned by \qrhead{} to each span at every decoding step.
Our analysis focuses on two metrics:
1) the fraction of decoding steps for which the gold edge (i.e., the next edge on the correct path) appears among the top 5\% most-attended spans (Gold Edge in Top 5\%). 2)  the total attention mass assigned to the gold edge at each decoding step (Gold Attention).
For this analysis (Figure~\ref{fig:motivation}, middle and right), we only consider the second step (edge) among the four steps to explicitly study the attention dynamics occurring in the middle of the reasoning process.

We additionally include randomly selected attention heads as a baseline for comparison.
As shown in Figure~\ref{fig:motivation} (middle), \textbf{\rethead{}s remain consistently aligned with the gold edge.}
Even as single-step prediction accuracy drops from over 90\% to approximately 20\% with increasing context length, \rethead{} continue to rank the gold edge among the top 5\% at most steps.
At the same time, \textbf{we observe a sharp decline in absolute attention to the gold edge}, closely mirroring the overall performance degradation (see Figure~\ref{fig:motivation} right).  
In contrast, random attention heads exhibit a much steeper drop in relative attention and much lower attention mass on gold spans.

\paragraph{Steering attention with retrieval heads.}
Our analysis find that \qrhead{} consistently ranks gold edges highly and preserves strong relative retrieval signals. However, as context length grows, absolute attention to gold spans declines across all heads, leading to performance degradation. This suggests that while \qrhead{} identifies relevant context, its signal is diluted.
Motivated by this observation, we investigate whether the stable retrieval behavior of \qrhead{} can be leveraged to steer overall attention toward relevant context at decoding time.

\section{\methodname{}: Dynamic Attention Scaling}

\paragraph{Overview}
The core idea of \methodname{} is to dynamically \emph{up-weight} tokens in the context based on the attention distribution of retrieval heads during decoding. As shown in Figure~\ref{fig:algo_overview}, each decoding step of \methodname{} contains three stages:
1) \textbf{Aggregation}: we run a partial forward pass over the input sequence to obtain the attention scores of \qrhead{} and use them to compute context relevance scores for the current generation step;
2) \textbf{Selection}: we select the most important tokens based on the relevance scores;
3) \textbf{Rescaling}: we run a full forward pass in which we up-weight the attention logits of selected tokens across \emph{all} attention heads.

\methodname{} only modifies the decoding procedure and requires no training. It relies solely on attention scores produced by the model itself. These properties make \methodname{} highly flexible: it can be directly applied to off-the-shelf LMs without any architectural changes, and it is compatible with arbitrary inputs and tasks without task-specific preprocessing.

\subsection{The \methodname{} Algorithm}
\label{sec:algo}
Algorithm~(\ref{alg:qrdecode}) details the full procedure of \methodname{}. Given an LM $\mathcal{M}$ with attention heads $\mathcal{H}$ and a set of retrieval  heads (\qrhead{}) $\mathcal{H}^{*}\subset \mathcal{H}$, we decode from an input sequence $\boldsymbol{x}_{\le T}$. At each decoding step $t \geq T+1$, \methodname{} maintains context relevance scores $\boldsymbol{r}=r_1,...,r_{t}$ where each $r_i$ denotes the relevance of each token $x_i$ in the prefix $\boldsymbol{x}_{\le t}$ for the current generation step. For clarity of presentation, we assume the relevance scores are properly initialized (line 2), and defer the details of initialization after we introduce the aggregation step. \methodname{} produces the next token in the following three stages:

\begin{wrapfigure}{r}{0.5\textwidth}
\vspace{-2em}
\begin{minipage}{0.475\textwidth}
\begin{algorithm}[H]

\fontsize{7.75}{7.75}\selectfont
\caption{\methodname{}}
\label{alg:qrdecode}

\textbf{Input:} LM $\mathcal{M}$ with attention heads $\mathcal{H}$, \qrhead{} heads $\mathcal{H}^{*}$, input $\boldsymbol{x}_{\le T}$, rescale strength $\beta$, momentum $\gamma$, cumulative probability threshold $p$ (for token selection).

\textbf{Output:} Generated sequence $\boldsymbol{x}_{\le T'}$
\begin{algorithmic}[1]
\STATE $t \gets T$
\STATE $\mathbf{r}_{T} \gets \mathbf{r}_{\mathrm{init}}$ \hspace{2em}\texttt{// Initialize Relevance}

\WHILE{not finished}
    \STATE $t \gets t + 1$

    \STATE \textbf{(Aggregation)} Run a partial forward pass to obtain attention logits from $\mathcal{H}^{*}$: $\mathbf{a}_t^{(h)}$ for $h \in \mathcal{H}^{*}$.

    \STATE $\mathbf{r}_t \gets \frac{1}{|\mathcal{H}^{*}|} \sum_{h \in \mathcal{H}^{*}} \mathrm{Softmax}(\mathbf{a}_t^{(h)})$
    \STATE $\mathbf{r}_t \gets \gamma \cdot \mathbf{r}_{t-1} + (1-\gamma)\cdot \mathbf{r}_t$ \hspace{2em}\texttt{// Apply Momentum}
    \STATE \textbf{(Selection)} $\boldsymbol{x}^{\text{*}} \gets \mathrm{SelectTop}
    (\boldsymbol{x}_{\le t}, \mathbf{r}_t,p)$ \hspace{2em}\texttt{// Select top-$p$ tokens}
    \STATE $\mathbf{v}[i] \gets \begin{cases}
        \log(\beta) & x_{i} \in \boldsymbol{x}^{*} \\
        0 & \text{otherwise}
    \end{cases}$ \hspace{1em}\texttt{//Intervention}

    \STATE \textbf{(Rescaling)} $\tilde{\boldsymbol{\alpha}}_t^{(h)} \gets \mathrm{Softmax}\!\left(\mathbf{a}_t^{(h)} + \mathbf{v}\right)$ for $h \in \mathcal{H}$
    \STATE $\ell_t \gets \mathcal{M}(\boldsymbol{x}_{\le t} \mid \tilde{\boldsymbol{\alpha}}_t)$ \hspace{2em}\texttt{// Rescaled Forward}
    \STATE $x_{t+1} \sim \mathrm{Softmax}(\ell_t )$ \hspace{2em}\texttt{// Sampling next token}
\ENDWHILE
\end{algorithmic}
\end{algorithm}
\end{minipage}
\vspace{-2em}
\end{wrapfigure}

\paragraph{Aggregation.}
We first perform a \emph{partial} forward pass at decoding step $t$, and extract the attention distributions of the selected \qrhead{} heads. The forward pass is partial as we skip the forward pass over higher layers given that retrieval heads are primarily located in the middle layers (see \S\ref{sec:algo_impl} for details).
We then compute a \emph{relevance score} over tokens in the prefix by \textit{averaging} the attention scores across these heads: $\mathbf{r}_t \;=\; \frac{1}{|\mathcal{H}^{*}|} \sum_{h \in \mathcal{H}^{*}} \boldsymbol{\alpha}^{(h)}_t$.

where  $\boldsymbol{\alpha}^{(h)}_t$ is the attention distribution after softmax at step $t$. To incorporate information from previous decoding steps, we maintain a moving average of the relevance scores with momentum $\gamma \in [0,1)$: $
\mathbf{r}_t \;\leftarrow\; \gamma \cdot \mathbf{r}_{t-1} + (1-\gamma) \cdot \mathbf{r}_t .$
\noindent Empirically, this momentum-based smoothing stabilizes the relevance estimates and makes \methodname{} more robust to hyperparameter variations. 

We note that at the first decoding step, we need to obtain an initial relevance score. We use a short warm-up window of length $T_w$, relying on the last $T_w$ tokens of the input prompt to initialize the relevance distribution.
Specifically, for the first decoding step $T$, we compute:

\footnotesize
$$\mathbf{r}_T
\;=\;
\mathrm{Normalize}\!\left(
\sum_{d=0}^{T_w-1} \gamma^{d} \cdot \boldsymbol{\alpha}^{\mathcal{H}^*}_{T-d}\right),$$
\normalsize
where $\boldsymbol{\alpha}^{\mathcal{H}^*}_{T-d}=  \frac{1}{|\mathcal{H}^{*}|} \sum_{h \in \mathcal{H}^{*}} \boldsymbol{\alpha}^{(h)}_{T-d}$ is the averaged attention of \qrhead{} at time $T-d$, and $\mathrm{Normalize}(\cdot)$ rescales the scores to form a valid distribution. We \emph{do not} tune $\gamma$ or $T_w$, and fix them to $\gamma=0.4$ and $T_w=8$ in all experiments.

\paragraph{Selection.}
At each decoding step $t$, we use the relevance distribution to estimate the importance of context tokens in the prefix $\boldsymbol{x}_{\le t}$. Based on this distribution, we select a subset of relevant tokens
 $\boldsymbol{x}^* \in \boldsymbol{x}_{\le t}$. Our selection strategy follows nucleus (top-$p$) sampling~\citep{Holtzman2020The} used for LM decoding, but is applied over relevance scores rather than next-token probabilities. Specifically, we rank tokens in $\boldsymbol{x}_{\le t}$ by their relevance scores and retain the smallest set of tokens whose cumulative relevance mass exceeds a threshold $p$, while additionally enforcing a maximum of $K$ (8192) tokens to avoid overly large number of selected tokens.
 
 In practice, we find performance to be robust to moderate variations in parameters. In all experiments, we determine these parameters using a validation set of fully synthetic tasks, and directly use the resulting parameters on diverse downstream datasets without further tuning (\S\ref{sec:experiments}). Concretely, we set top-$p$ to be 0.95 or 0.975.

\paragraph{Rescaling.}
Given the selected relevant tokens $\boldsymbol{x}^{*}$, we intervene on the attention computation to up-weight the selected tokens by modifying the attention logits. Concretely, for each layer and each attention head, we compute an intervention vector of bias terms added to the attention logits. The bias is $\log \beta$ for the selected tokens, and $0$ otherwise, where $\beta > 1$ is the rescale strength factor: $\mathbf{v}[i] = \log(\beta)\;\; \text{if } x_i \in \boldsymbol{x}^*, \text{ else } 0.$

Then, we add the intervention vector to the attention logits $\mathbf{a}_t^{(h)}$ of \emph{all} attention heads $h \in \mathcal{H}$ before the softmax operation, and obtain rescaled attention distributions $\tilde{\boldsymbol{\alpha}}_t^{(h)}$.
We apply the intervention to all heads to propagate the relevance signal identified by \qrhead{} throughout the model’s attention computation.
{\footnotesize\[
\tilde{\boldsymbol{\alpha}}_t^{(h)} = \mathrm{Softmax}\!\left(\mathbf{a}_t^{(h)} + \mathbf{v}\right) \text{ for } h \in \mathcal{H}
\]}
This intervention effectively rescales the pre-softmax attention logits by $\beta$.
Finally, we perform a full forward pass with the modified attention logits across all layers, and the next token $x_{t+1}$ is sampled or selected from the output distribution produced by this rescaled forward pass. Importantly, this procedure operates entirely at inference time and does not require any modification to model parameters.

\subsection{Efficient Implementation}
\label{sec:algo_impl}

\paragraph{Early stopping of attention aggregation.}
We describe our design that makes \methodname{} more efficient. Prior work shows that attention heads responsible for retrieval and reasoning tend to concentrate in the middle layers of transformer models \citep{wu2025retrieval,zhao2024understandingsyntheticcontextextension}.
For instance, for Qwen3-8B (36 layers), \qrhead{} are distributed between 17$-$20 layers. Hence, we early-stop attention aggregation during the partial forward pass and only collect \qrhead{} attention up to the middle layers of the model, rather than across all layers. With early stopping, this adds approximately $60\%$ extra computation \emph{per decoding step}.

\paragraph{Overhead analysis.} As mentioned above, the attention aggregation stage introduces an additional \emph{partial} forward pass during decoding, while leaving prefilling unchanged.
In general, \methodname{} incurs a small FLOP overhead (e.g., $\sim$6\% when generating 8K tokens with a 128K context), no additional peak memory cost, and a reduction in throughput due to extra decoding-time computation.
We provide a detailed analysis in Appendix~\ref{app:effeciency_discussion}.

\section{Experiments}
\label{sec:experiments}
We test the effectiveness of \methodname{} on a diverse array of long-context reasoning tasks.

\paragraph{Models.}
We experiment with multiple open-weight LMs from two families, including \textbf{Qwen3-4B, 8B, and 32B} of Qwen3 family~\citep{Yang2025Qwen3TR}, and \textbf{Llama-3.1-8B-Instruct}~\citep{dubey2024llama}. Qwen3 models support thinking mode that uses long CoT~\citep{guo2025deepseek}, and Llama-3-8B-Instruct can use CoT prompting~\citep{chain}.

\paragraph{Head selection.} We directly use the set of \qrhead{}~\citep{qrheads} from the official implementation across all experiments.
We note that these heads are detected on Natural Questions~\citep{natq} and applied directly across all tasks evaluated in this paper, demonstrating strong cross-task transfer.

\paragraph{Baselines.}
Our approach operates on the decoding process for improving long-context reasoning. We compare against baselines that directly operate on decoding, including:

1) \textbf{Vanilla}, standard decoding without alteration.

2) \textbf{YaRN}~\citep{peng2024yarn}, which extends context window by modifying rotary position embeddings~\citep{su2024roformer}. YaRN primarily targets extending the usable context window of pretrained models, rather than improving long-context reasoning accuracy.

3) \textbf{Uniform Attention Scaling (\attnts{})}.
Prior work has shown that sharpening attention distributions can improve long-context extrapolation~\citep{peng2024yarn,chen2026critical,Nakanishi2025ScalableSoftmaxIS}. Concretely, this is achieved by applying a temperature $\tau \in (0,1]$ to rescale attention logits, i.e., $\mathbf{a}' = \mathbf{a}/\tau$, where $\mathbf{a}$ denotes the original attention logits.
Following this line of work, we adopt length-dependent attention temperature scaling, using different values of $\tau$ for different input lengths. While prior approaches typically scale attention with a factor that grows linearly with $\log n$, we instead tune $\tau$ separately for each context length.
Importantly, \attnts{} rescales attention logits \emph{uniformly} across all tokens, whereas \methodname{} selectively up-weights step-relevant tokens at each decoding step.

Both \attnts{} and \methodname{} can be combined with YaRN.

\paragraph{Setting parameters for \methodname{} and \attnts{}.} \methodname{} and \attnts{} both require setting additional parameters for controlling the decoding. We use \mrcr{}~\citep{mrcr} (Multi-Round Coreference Resolution) as a development dataset for deciding hyperparameters. We use MRCR as it is a fully synthetically generated dataset, which makes it less biased for a particular real-world dataset. For each model, we pick one set of parameters for length span (0,64K) and one set of parameters for length span (64K, 128K) using performance on MRCR, and we directly apply the \emph{same set of parameters} on other datasets.
We include more details in Appendix~\ref{app:detail_param_selection}.

\begin{wrapfigure}{r}{0.49\linewidth}
\vspace{-1.5em}
\centering
\footnotesize
\fontsize{7.75}{7.75}\selectfont
\begin{tabular}{lcccc}
\toprule
\bf \pathtrans{} & \bf 4K & \bf 8K & \bf 16K & \bf 32K \\
\midrule
Qwen3-32B & 80 & 67 & 32 & 21 \\
\, + \attnts & \textbf{82} & 68 & 40 & 19 \\
\, + \methodname & 79 & \textbf{71} & \textbf{46} & \textbf{33} \\
\midrule
Qwen3-8B & 57 & 45 & 19 & 1 \\
\, + \attnts{} & 59 & 46 & 25 & 2 \\
\, + \methodname & \textbf{65} & \textbf{51} & \textbf{33} & \textbf{2} \\
\midrule
\end{tabular}
\captionof{table}{Performance across different context lengths with Uniform Attention Scaling (\attnts{}) and \methodname{}.}
\label{tab:check_path_trans}
\end{wrapfigure}

\paragraph{Diagnostic Test on \pathtrans{}}
We first use \pathtrans{} to evaluate whether reshaping attention with \methodname{} or \attnts{} helps LMs maintain focus on relevant context at each decoding step.
We do not include YaRN in this setting, as the maximum input length (32K) is well within the supported context window of the evaluated models. Results are reported in Table~\ref{tab:check_path_trans}.
As shown, \methodname{} yields consistent performance improvements across context lengths.
\attnts{} also leads to measurable gains for Qwen3-8B at 4K and 32K input lengths, though the improvements are smaller and less consistent.
These results suggest that attention sharpening alone can yield modest improvements, whereas explicitly up-weighting important tokens identified by \qrhead{} leads to substantially larger gains.

\subsection{Long Context Reasoning Tasks}
\label{sec:exp_cot}
We now evaluate \methodname{} on the primary focus of this paper, improving multi-step CoT reasoning over long context.

\paragraph{Datasets and settings.} We briefly summarize the datasets and settings used for our experiments. Please refer to Appendix~\ref{app:detail_exp_setup} for more details.

We use three datasets with natural texts: 1) \textbf{\mrcr{}}~\citep{mrcr} requires LMs to retrieve a conversation following a query from a list of highly similar conversations. 2) \textbf{\longbench{}}~\citep{longbenchv2} requires multi-step reasoning over realistic long contexts spanning diverse domains. 3) \textbf{\textsc{Clipper}}~\citep{clipper} evaluates claim verification over full-length books (90–128K tokens), requiring multi-step reasoning over evidence distributed throughout the text.

For Qwen3 models, we activate thinking mode to use long CoT (4096-token budget for \mrcr{}, \clipper{} and 10{,}240-token for \longbench{}). For 128K-context experiments, we use YaRN (factor 4.0) for Qwen3 models to extend their context window from 32K to 128K, following the recommendation in \citet{Yang2025Qwen3TR}. We note that both \attnts{} and \methodname{} are applied on top of YaRN for 128K. For Llama model, we use CoT prompting template from original datasets to enable CoT. 
We do not evaluate Llama models in this setting, as they do not support long CoT (direct recall results are reported  in \S\ref{sec:exp_long_recall}).

\begin{wrapfigure}{r}{0.525\textwidth}
    \centering
    \includegraphics[width=\linewidth]{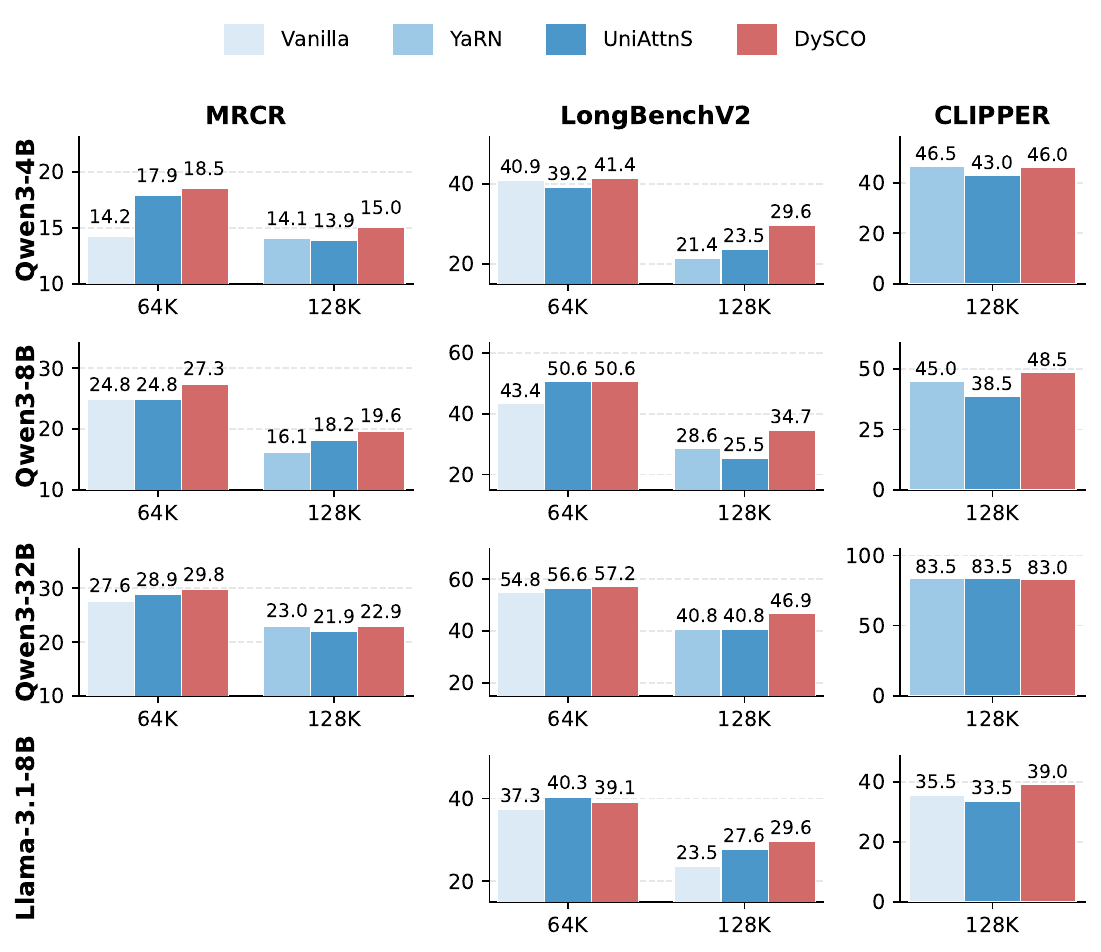}
    \vspace{-1em}
    \caption{Performance on \mrcr{}, \longbench{}, and \clipper{}. \methodname{} substantially outperforms vanilla decoding, and \attnts{}. YaRN is applied to Qwen models at 128K context length, but not to Llama-3.1-8B-Instruct, which natively supports 128K.
}
\vspace{-2em}
    \label{fig:main_cot}
\end{wrapfigure}

\paragraph{Results.}

As shown in Figure~\ref{fig:main_cot}, \textbf{\methodname{} improves CoT reasoning under long-context settings} across model families, benchmarks, and context lengths. At 128K context length, \methodname{} (with YaRN) improves Qwen3-8B by 3.5 absolute points (22\% relative) on \mrcr{}, 6.7 points (18\%) on \longbench{}, and \textbf{3.5} points on \clipper{} compared to YaRN alone.  While \attnts{} yields gains in some cases (e.g., a 7.2\% improvement for Qwen3-8B on \longbench{} at 64K), these improvements are not consistent across datasets or context lengths. In contrast, \methodname{} provides larger and more stable gains, particularly at longer context lengths, and consistently outperforms \attnts{} across all benchmarks.

\subsection{Long Context Recall Tasks}
\label{sec:exp_long_recall}

We next evaluate \methodname{} on long-context recall tasks, which primarily follow a needle-in-the-haystack setup~\citep{niah}. For these tasks, models directly produce the final answer without using CoT (we turn off thinking for Qwen models).

\paragraph{Datasets.}  We evaluate \mrcr{} under its intended direct-answering setup (directly outputting the final response). Additionally, we consider the following tasks: HotpotQA~\citep{hotpot}, QA and multiple-choice tasks from InfBench~\citep{infbench}. For Hotpot and InfBench, we use the processed versions released in HELMET~\citep{helmet}, and evaluate all tasks with a context length setting of 128K tokens.

\begin{wrapfigure}{r}{0.49\linewidth}
\vspace{-1em}
\centering
\renewcommand{\tabcolsep}{1.0mm}
\footnotesize
\fontsize{7.}{7.}\selectfont
\begin{tabular}{lccccc}
\toprule
 & \mrcr{} & \hotpot{} & \infmc{} & \infqa{} & Avg \\
 & 128K & 128K & 128K & 128K & 128K \\
\midrule

\textbf{Qwen3-4B\textsuperscript{YaRN}} & 16.7 & 37 & 50 & 27.1 & 32.7 \\
+ \attnts{} & 18.4 & 34 & 39 & 24.1 & 28.9 \\
+ \methodname{} & \textbf{19.0} & \textbf{39} & \textbf{54} & \textbf{31.3} & \textbf{35.8} \\
\midrule
\textbf{Qwen3-8B\textsuperscript{YaRN}}& 16.7 & \textbf{61} & 61 & \textbf{39.6} & 44.6 \\
+ \attnts{} & 17.5 & 58 & 63 & 39.7 & 44.6 \\
+ \methodname{} & \textbf{19.5} & 58 & \textbf{65} & 39.2 & \textbf{45.4} \\
\midrule
\textbf{Qwen3-32B\textsuperscript{YaRN}} & 24.4 & 51 & \textbf{79} & \textbf{48.7} & 50.8 \\
+ \attnts{} & 24.2 & 51 & 78 & 48.6 & 50.5 \\
+ \methodname{} & \textbf{25.3} & \textbf{53} & 77 & \textbf{48.7} & \textbf{51.0} \\
\midrule
\textbf{Llama3.1-8B-Inst.} & 17.2 & 46 & 61 & 37.1 & 40.3 \\
+ \attnts{} & \textbf{18.9} & 51 & 56 & 35.7 & 40.4 \\
+ \methodname{} & 18.6 & \textbf{52} & \textbf{62} & \textbf{38.4} & \textbf{42.8} \\
\bottomrule
\end{tabular}
\captionof{table}{Performance on long-context recall tasks with 128K input length. We apply YaRN for Qwen models.}
\label{tab:main_direct}
\vspace{-2em}
\end{wrapfigure}

\paragraph{Results.}
As shown in Table~\ref{tab:main_direct}, \methodname{} also able to improve model accuracy on long-context recall tasks. For example, with Llama-3.1-8B-Instruct, \methodname{} increases HotpotQA accuracy from 46 to 52, and InfQA accuracy from 37.1 to 38.4. While \attnts{} yields modest gains on MRCR, it also degrades performance in several other cases. Notably, these recall tasks require only very short outputs (typically tens of tokens), and \methodname{} introduces negligible additional overhead compared to vanilla decoding.

\subsection{Ablations \& Analysis}
\label{sec:analysis}

\paragraph{Ablation: Importance of head selection.}
We compare \methodname{} instantiated with \qrhead{} against a variant that uses randomly selected heads. Both variants use 16 heads. Table~\ref{tab:ablation} reports results on multiple datasets with Qwen3-8B. \methodname{} with random heads can outperform vanilla decoding on some tasks, since random heads may still capture weak retrieval signals (\S\ref{sec:qrhead_analysis}). Using \qrhead{} consistently yields the best results across all settings, highlighting the advantage of more reliable relevance signals.

\begin{wrapfigure}{r}{0.48\textwidth}
\centering
\footnotesize
\fontsize{7.9}{7.9}\selectfont
\renewcommand{\tabcolsep}{0.7mm}
\begin{tabular}{lccccc}
\toprule
 \bf  & \multicolumn{2}{c}{\bf MRCR } &  \bf LongV2 & \bf Path& \bf InfMC \\
 \bf  & \multicolumn{2}{c}{\bf 64K } &  \bf 64K & \bf 16K& \bf 128K \\
 
Think           & Yes & No & Yes & No & No  \\
\midrule
Vanilla & 24.8 & 24.8 & 43.4 & 19 & 61 \\
\methodname{} & \textbf{27.3} & \textbf{26.3} & \textbf{51.2} & \textbf{33} & \textbf{66} \\
w/ RandomHead & 26.6 & 24.2 & 47.0 & 20 & 63 \\
w/ StaticScaling & 22.5 & 23.1 & 47.5 & 23 & 64 \\
\bottomrule
\end{tabular}
\captionof{table}{Ablations of \methodname{} on head selection and dynamic rescaling with Qwen3-8B. \methodname{} outperforms its ablations with random attention heads and with static scaling.}
\label{tab:ablation}

\end{wrapfigure}

\paragraph{Ablation: Importance of dynamic rescaling.} 
We further ablate the role of dynamic rescaling by comparing \methodname{} with a static rescaling variant. Static rescaling applies attention reweighting to a \emph{fixed} set of context tokens, selected during an initial warm-up stage, and reuses this set throughout decoding. Conceptually, static rescaling can be viewed as an adaptation of KV-cache eviction methods that avoids explicit context pruning, as it relies on similar criteria for identifying important tokens (e.g., SnapKV). As shown in Table~\ref{tab:ablation}, static rescaling yields improvements on recall tasks (e.g., InfMC), but it consistently underperforms dynamic rescaling. This gap highlights the importance of dynamic rescaling in \methodname{}, as the set of relevant tokens can shift substantially over the course of generation.

\paragraph{Analysis: Hyperparameter robustness.} We examine the robustness of \methodname{} to variations of its hyperparameters. \methodname{} shows consistent performance across reasonable choices of hyperparameters. Full results are provided in Appendix~\ref{app:param_robustness}.

\paragraph{Analysis: Impact of context length.}
We analyze how the effectiveness of \methodname{} varies with context length by comparing results from \S\ref{sec:exp_cot} and \S\ref{sec:exp_long_recall}. The gains are particularly pronounced for CoT reasoning under long-context settings. See Appendix~\ref{app:cot_interplay} for details.

\paragraph{Additional comparisons.}
We provide further comparisons in Appendix~\ref{app:comp_rag}, showing that \methodname{} outperforms retrieval-augmented generation and prompt compression approaches.

\paragraph{Qualitative examples.} We provide qualitative examples showcasing how \methodname{}  effectively steers LMs in long-context reasoning in Appendix~\ref{app:qualitative_example}. 

\paragraph{Discussion: scope of our approach.} \methodname{} is designed for long-context regimes where the input context is substantially longer than output. As an inference-time technique, \methodname{} can be applied selectively at deployment: it can be enabled for long inputs, and disabled for shorter inputs. This flexibility allows practitioners to adapt the method based input length, since its benefits are most pronounced in long-context settings rather than universally across all generation regimes.

\section{Related Work}

We focus our discussion on closely related inference-time techniques for long-context LMs. Please refer to Appendix~\ref{app:extended_related_work} for more general methods for long-context modeling.

\paragraph{Efficient long-context inference techniques.}
One line of work studies inference-time techniques for long context, primarily focusing on improving efficiency. Representative approaches include streaming inference~\citep{xiao2024efficient,zhang2023h2o}, KV cache eviction or compression~\citep{Xu2024RecycledAE,li2024snapkv,cai2024pyramidkv,guilio2024finch,kim2024infinipot}, and sparse attention mechanisms~\citep{xu2025ftp,jiang2024minference,xiao2025duoattention}.  
These methods trade off accuracy for computational or memory efficiency. In contrast, \methodname{} is explicitly designed to improve \emph{accuracy} under long context.

\paragraph{Inference techniques for long-context modeling.}
Recent work has explored improving long-context modeling at inference time, including training on input context at test time~\citep{Bansal2025LetsJ,Chen2025PERKLR} and learning intervention vectors over query and key matrices~\citep{Zhu2025FocusDM}. These approaches typically require additional training or parameter updates. In contrast, \methodname{} requires no training and can be directly applied to off-the-shelf LMs.
Most closely related to our approach, prior work has explored modifying attention distributions to better utilize long context. This includes global attention scaling methods that apply a uniform rescaling factor to attention logits~\citep{peng2024yarn,Nakanishi2025ScalableSoftmaxIS,chen2026critical,puvvada-etal-2025-swan}, as well as positional scaling strategies that apply fixed, position-dependent adjustments to mitigate the lost-in-the-middle problem~\citep{liu-etal-2024-lost,hsiehfound,zhang2024found}. In contrast, \methodname{} performs selective rescaling and up-weights attention to task-relevant tokens, enabling more targeted and effective improvements in long-context reasoning.

\paragraph{Scaffolding and external systems for long context.}
Another line of work builds external scaffolding around LMs. This includes RAG~\citep{zhao-etal-2024-longrag,li-etal-2024-retrieval}, prompt compression modules~\citep{longllmlingua,Wu2025ReSumUL}, recursive or multi-stage LM calls~\citep{zhang2025recursive}, memory systems~\citep{chen2024walking}, and agentic frameworks~\citep{zhang2024chain,zhao-etal-2024-longagent}. 
\methodname{} directly improves the LM's internal attention behavior during decoding, without introducing additional scaffolding, and can be integrated with any scaffolding.

\section{Conclusion}

We have presented \methodname{}, a training-free decoding algorithm that improves long-context reasoning. \methodname{} leverages retrieval heads to dynamically identify relevant context tokens at each decoding step and intervenes on attention logits to up-weight their attention across heads. Across multiple instruct and reasoning LMs, \methodname{} consistently improves accuracy on challenging long-context benchmarks while incurring modest extra FLOPs. Our analyses further indicate that long-context failures are associated with degraded focus on relevant context, and that dynamic attention scaling can mitigate this issue. Overall, \methodname{} provides an inference-time approach for improving long-context accuracy and offers insight into the mechanisms underlying long-context failures in LMs.

\bibliography{colm2026}
\bibliographystyle{colm2026_conference}

\newpage
\appendix

\section{Additional Details on Query-Focused Retrieval Heads (\qrhead{}).} 
\label{app:detail_qrhead}
Original retrieval heads are identified through strict copy behavior, which overlooks more general semantic retrieval. To address this limitation, \citet{qrheads} propose Query-Focused Retrieval Heads (QRHeads), which generalize retrieval behavior to query-conditioned context lookup.

Specifically, consider a long-context QA setting in which the input consists of a query $q$ and a large context containing one gold document $d^*$ (the ``needle'') along with many distractor documents. \citet{qrheads} define a query-context retrieval score (QRScore) for each attention head $h$ as:

\[
\mathrm{QRScore}(h) = \sum_{i \in q} \sum_{j \in d^*} \alpha^{(h)}_{i,j},
\]

\noindent where $\alpha^{(h)}_{i,j}$ denotes the attention weight from query token $i$ to context token $j$ under head $h$. Heads are ranked by their QRScore, and the top-$K$ heads (typically $1$--$2\%$ of all attention heads) are selected as QRHeads.

Unlike original retrieval heads, QRHeads capture semantic context lookup rather than exact token copying. Prior work shows that attention from QRHeads is more effective for retrieving relevant information from long contexts across tasks and domains~\citep{qrheads}.

\section{Extended Related Work}
\label{app:extended_related_work}

In the main body of the paper, we focus on closely related inference-time techniques; here, we provide a overview of work on improving long-context LMs more broadly.

The need for better support of long context has motivated extensive research across multiple dimensions, including architectural designs~\citep{gu2024mamba,Lieber2024JambaAH,peng-etal-2023-rwkv,Xiao2023EfficientSL,bertsch2023unlimiformer,jin2024llm,yen-etal-2024-long}, data engineering strategies~\citep{xiong2023effective,an2024make,prolong,hu2024longrecipe,fudata,chen2025longpo,gao2025nextlong,bai2024longwriter,chen-etal-2025-ladm}, context window extension techniques~\citep{peng2024yarn,chen2024longlora,chen2023extending,zhu2024pose}, evaluation benchmarks~\citep{helmet,Longproc,bai2023longbench,longbenchv2,gsmlonggen,periodiclonggen,hsieh2024ruler,novelqa}, and analyses of long-context behavior and failure modes~\citep{wu2025retrieval,zhao2024understandingsyntheticcontextextension,reallyhardlongcontext,Liu2023LostIT,gao-etal-2024-insights,yin2024lofit}.
Most existing approaches improve long-context performance by modifying model parameters, architectures, or training data, whereas our work improves the decoding procedure. 
In contrast, our work explores a new direction: improving long-context accuracy purely at inference time by modifying the decoding procedure.

\section{Details of Experimental Setup}
\label{app:detail_exp_setup}

\paragraph{Datasets.} We use 3 datasets with natural texts and enable CoT as follows:

1) \textbf{\mrcr{}}~\citep{mrcr} requires LMs to retrieve a conversation following a query from a list of highly similar conversations. The task is introduced to test LMs' direct recall. To allow LMs to use CoT, we activate thinking mode (4096-token budget) for Qwen3 models. We do not evaluate Llama models in this setting, as they do not support long CoT (direct recall results are reported  in \S\ref{sec:exp_long_recall}).

2) \textbf{\longbench{}}~\citep{longbenchv2} requires multi-step reasoning over realistic long contexts spanning diverse domains. We use a subset of data points with input length ranging from 0-128K (264 data points). For Qwen models, we enable thinking mode with a maximum of 10{,}240 tokens to accommodate the difficulty; for Llama models, we apply CoT prompting from \citet{longbenchv2}.

3) \textbf{\textsc{Clipper}}~\citep{clipper} evaluates claim verification over full-length books. The model is given a complete book (90–128K tokens) along with a claim to verify, requiring multi-step reasoning over evidence distributed throughout the text. We include this dataset because it provides a native CoT prompting template, which is well-suited for evaluating instruction-tuned models (Llama-3.1-8B-Instruct). We follow the original experimental setup and response template from \citet{clipper}.

\paragraph{Settings.} We use YaRN (factor 4.0) for Qwen3 models to extend their context window from 32K to 128K, following the recommendation in \citet{Yang2025Qwen3TR}. For 128K-context experiments, both \attnts{} and \methodname{} are applied on top of YaRN. For Llama models with explicit CoT prompting, we use greedy decoding.  For Qwen3 models, we use the recommended generation configurations, while fixing the random seed for decoding across different methods per run, following ~\citet{Bansal2025LetsJ,longbenchv2} to control the variance across methods.

\section{Detailed Discussion of Overheads}
\label{app:effeciency_discussion}

We begin by briefly reviewing the standard generation procedure of language models. Given an input prefix $x_{<T}$, the model generates outputs $x_{T:T'}$ autoregressively until termination. This process consists of two phases: (1) \emph{prefilling}, where the full input $x_{<T}$ is processed in a single forward pass to construct KV caches, and (2) \emph{autoregressive decoding}, where tokens are generated one at a time conditioned on the cached states.

\methodname{} operates only at the autoregressive decoding phase, while the prefilling stage remains unchanged and can leverage standard optimizations such as Flash Attention~\cite{dao2022flashattention,dao2023flashattention2}. This design ensures that \methodname{} introduces no additional peak memory overhead from prefilling, and isolates all extra cost to decoding-time computation. We analyze this overhead in terms of FLOPs, memory, and throughput below.

\paragraph{FLOPs.}
As mentioned in \S\ref{sec:algo_impl}, the attention aggregation stage introduces an additional \emph{partial} forward pass during decoding, but no additional cost during prefilling. As a result, the total FLOP overhead of \methodname{} remains small for long-context workloads, where computation is dominated by the quadratic-cost prefilling pass over the input context. To make this concrete, consider a setting with a 100K-token input context. Following ~\citet{Hoffmann2022TrainingCL}, our estimate of the FLOPs is that generating 4K and 8K output tokens accounts for roughly $4\%$ and $8.4\%$ of the prefilling FLOPs, respectively. When using \methodname{}, the additional partial decoding pass increases the total computation by only about $2.5\%$ and $5.0\%$ relative to prefilling for these two settings.

\begin{table}[h]
\centering
\renewcommand{\tabcolsep}{0.7mm}
\begin{tabular}{lcc}
\toprule
 & EAGER (Implementation) & Flash (Theoretical) \\
\midrule
Vanilla & 6.1 & 17.8 \\
\methodname{} & 3.6 & 10.5 \\
\bottomrule
\end{tabular}
\caption{Throughput (tokens/s) comparison between vanilla decoding and \methodname{}, under standard attention (EAGER) and an estimated Flash Attention implementation. Results are obtained on Qwen3-8B with 100K input tokens and 4K output tokens on an H100 GPU.}
\label{tab:throughput}
\vspace{-1em}
\end{table}

\paragraph{Peak memory.}
\methodname{} does not increase peak memory usage. In standard decoding, peak memory typically occurs during prefilling, which processes the full input sequence in a single forward pass. In contrast, \methodname{} only materializes attention patterns for a small subset of heads during decoding (i.e., $|\mathcal{H}^*| \times 1 \times T$), rather than full attention maps across all heads and tokens.

As a result, the peak memory footprint remains dominated by prefilling. For example, with a 100K-token input and 4K-token output, prefilling requires approximately 39GB of memory, while decoding requires only $\sim$32GB. \methodname{} does not exceed this bound.

\paragraph{Throughput.}
\methodname{} introduces additional computation during decoding due to the partial forward pass, leading to a reduction in throughput. In our implementation, this results in an approximate $1.7\times$ slowdown compared to vanilla decoding. Table~\ref{tab:throughput} reports throughput for generating 4K tokens conditioned on 100K inputs.
This reflects our design that \methodname{} sacrifices decoding latency for improved long-context accuracy. We note that further optimizations could be possible. For example, applying interventions only from intermediate layers after identifying relevant signals from \qrhead{}, which we leave for future work.

\paragraph{Compatibility with Flash Attention.}
Our current implementation uses eager attention during decoding for simplicity, as we intervene directly on attention logits. However, the intervention in \methodname{} can be expressed as an additive bias to attention scores, which can in principle be incorporated into Flash Attention’s block-wise computation without materializing full attention matrices~\citep{dao2022flashattention}.
\methodname{} can be made compatible with Flash Attention with modest overhead, though we leave a full implementation to future work.

\section{Comparison with RAG and Prompt Compression}
\label{app:comp_rag}

\begin{wrapfigure}{r}{0.45\textwidth}
 \centering
        \includegraphics[width=\linewidth]{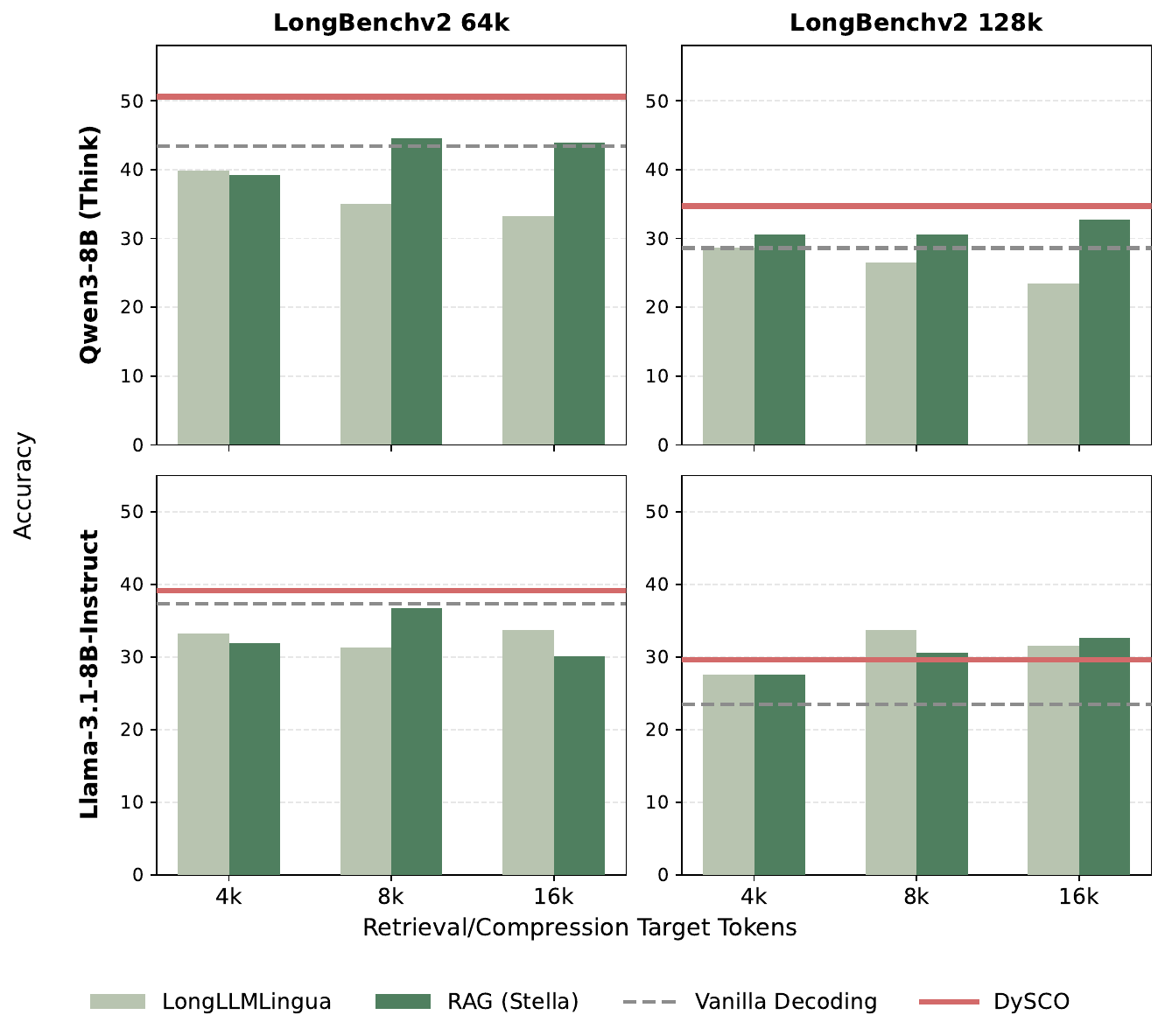}
        \caption{Comparison between \methodname{}, RAG (Stella), LongLLMLingua, and vanilla decoding. For RAG and LongLLMLingua, we report the results after reducing the context to different length (4K, 8K, and 16K tokens).}
        \label{fig:rag}
\end{wrapfigure}

we compare our decoding-time approach with methods that reduce the effective context length via external scaffolding. Specifically, we consider two representative classes of approaches:
(1) Retrieval-Augmented Generation (RAG), which employs a dense retriever to select the most relevant portions of the input context given a query; and
(2) Prompt compression methods, which explicitly rewrite or prune the input prompt to fit within a shorter token budget. In particular, we evaluate LongLLMLingua~\citep{longllmlingua}, which uses an auxiliary LM to compress long prompts while attempting to preserve task-relevant information.

Both RAG and prompt compression methods are less flexible than \methodname{}. They require additional system components (e.g., external retrievers or compression models) and rely on explicit partitioning of the input into instruction, context, and query segments. In contrast, \methodname{} operates purely at decoding time on the original input sequence, without modifying the prompt or introducing any external scaffolding.

We evaluate \methodname{}, RAG, and LongLLMLingua on the challenging long-context reasoning benchmark \longbench{}. We do not include benchmarks such as MRCR and InfBench, which are specifically designed to stress-test long-context processing and are largely solvable by RAG-style designs.

\paragraph{Setup.}
For RAG, we adopt a strong dense retriever, Stella-1.5B-V5~\citep{zhang2025stella}. Following \citet{longbenchv2}, we apply a chunk-and-concatenate strategy to reduce the context length. Specifically, we first partition the original long context into chunks of 1024 tokens. Given a query, Stella V5 retrieves the top-${4,8,16}$ most relevant chunks, which are then concatenated in their original order, resulting in shortened contexts of 4K, 8K, and 16K tokens. For LongLLMLingua, we use the official implementation released by \citet{longllmlingua}. We partition the input prompt into context and question components, set the target compressed length to ${4\text{K}, 8\text{K}, 16\text{K}}$ tokens, and feed the compressed prompt to the model for decoding.

\paragraph{Results.}
Figure~\ref{fig:rag} summarizes the results. Overall, \methodname{} outperforms both RAG and LongLingua on Qwen3-8B (Think), and achieves performance comparable to these baselines on Llama-3.1-8B. As both models have undergone substantial post-training and already exhibit strong long-context capabilities, neither RAG nor LongLingua yields consistent improvements on \longbench{} when the input context length is within 64K tokens. In contrast, \methodname{} continues to improve performance in this regime. When the context length increases to 128K tokens, RAG-based systems begin to outperform vanilla decoding. However, they still lag behind \methodname{} on Qwen3-8B.

\section{Impact of CoT Reasoning and Context Length}
\label{app:cot_interplay}

\begin{wrapfigure}{r}{0.45\textwidth}
\centering
        \includegraphics[width=\linewidth]{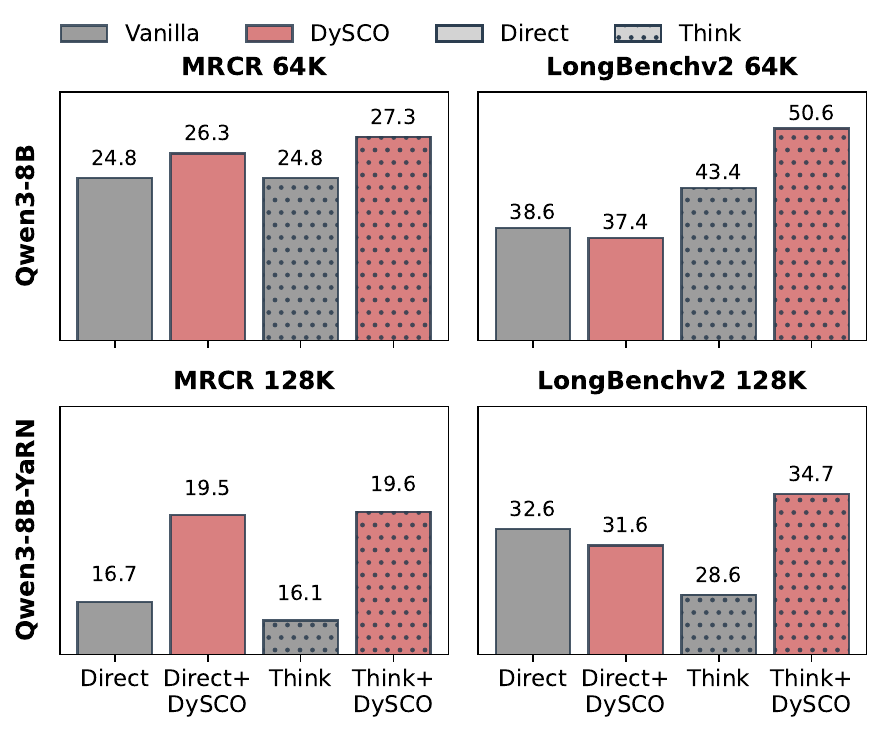}
        \caption{Performance of Qwen3-8B on \mrcr{} and \longbench{} at 64K and 128K under four decoding settings: direct answer vs. Think, with vanilla decoding or \methodname{}. \methodname{} yields larger gains when combined with CoT.
        }
        \label{fig:cot_interplay}
\end{wrapfigure}

\methodname{} dynamically focuses on key context as generation progresses, making it naturally compatible with CoT, where different reasoning steps rely on different parts of the context. We analyze the interaction between \methodname{} and CoT across varying context lengths. We evaluate \methodname{} and vanilla decoding with and without CoT on \mrcr{} and \longbench{} at 64K and 128K context lengths, under four decoding settings (direct answer vs. CoT, each with vanilla decoding or \methodname{}). We report the results for Qwen3-8B.

\paragraph{Interplay between \methodname{} and CoT reasoning.} Figure~\ref{fig:cot_interplay} summarizes the results. Overall, \textbf{\methodname{} yields larger benefits when combined with CoT reasoning}. On \mrcr{}, which primarily evaluates long-context recall, \methodname{} improves performance in both the direct-answer and CoT settings, with larger gains when CoT is enabled. On \longbench{}, which emphasizes multi-step reasoning over long contexts, \methodname{} substantially improves performance in the CoT setting, while causing only minor degradation when CoT is disabled. We hypothesize that this occurs because \longbench{} is inherently reasoning-intensive, and disabling CoT limits the model’s ability to capitalize on improved attention to relevant context.

\paragraph{Impact of context length.} We find that the interaction between CoT reasoning and decoding strategy depends on input length. 
Under vanilla decoding, CoT outperforms direct answering at 64K context length, but not at longer context lengths (128K). 
However, \methodname{} reverses this trend at 128K: \textbf{when combined with CoT reasoning, \methodname{} substantially improves performance at longer input lengths}, where direct-answer decoding is comparatively weaker.
By dynamically rescaling attention, \methodname{} restores effective long-context reasoning at 128K input length.

\section{Parameter Selection for \methodname{} and \attnts{}}
\label{app:detail_param_selection}

Our method performs inference-time attention scaling to improve long-context accuracy. Both \methodname{} and the baseline \attnts{} introduce a small number of inference-time hyperparameters. Here, we describe 1) how these parameters are selected, and 2) the robustness of \methodname{} to reasonable variations around the chosen defaults.

\subsection{Choosing Parameters for \methodname{}}
\label{app:param_qrdecode}

We select hyperparameters using \mrcr{}, a fully synthetic long-context recall benchmark. We choose \mrcr{} for parameter tuning for two main reasons: 1) it is synthetically generated and therefore does not risk data contamination with real-world evaluation benchmarks, and 2) its context length is configurable, allowing controlled analysis across different lengths. Importantly, for each model, we fix a single set of hyper-parameters and apply it uniformly across all downstream tasks.

For Qwen models, we distinguish between two context-length settings. For inputs up to $64$K tokens, we use the native context window without extrapolation. For $64$--$128$K tokens, we use YaRN-based extrapolation, which globally changes the attention computation. As a result, we select parameters separately for the $0$--$64$K and $64$--$128$K settings, using \mrcr{}-$64$K and \mrcr{}-$128$K, respectively. Within each length span, a single parameter configuration is shared across all tasks.

\paragraph{Hyperparameters.}
We perform a small grid search on \mrcr{} over the following values:
\[
p \in \{0.95, 0.975\}, \quad
K \in \{4096, 8192\}, \quad
\beta \in \{2.0, 2.5, 3.0\}.
\]
When multiple configurations yield comparable performance, we prefer less aggressive settings—specifically, larger $p$, smaller $\beta$, and smaller $K$—as they intervene more conservatively on the attention distribution.

The parameter choices used throughout the paper are summarized below:
\begin{center}
\footnotesize
\begin{tabular}{lccc}
\toprule
\textbf{Model / Length} & $\boldsymbol{p}$ & $\boldsymbol{K}$ & $\boldsymbol{\beta}$ \\
\midrule
Qwen3-4B (0--64K) & 0.975 & 4096 & 2.0 \\
Qwen3-4B + YaRN (64--128K) & 0.95 & 8192 & 2.5 \\
Qwen3-8B (0--64K) & 0.975 & 4096 & 2.0 \\
Qwen3-8B + YaRN (64--128K) & 0.975 & 8192 & 2.5 \\
Qwen3-32B (0--64K) & 0.975 & 4096 & 2.0 \\
Qwen3-32B + YaRN (64--128K) & 0.95 & 8192 & 2.5 \\
LLaMA-3.1-8B (0--128K) & 0.975 & 4096 & 2.0 \\
\bottomrule
\end{tabular}
\end{center}

We observe that models within the same family exhibit consistent optimal configurations. For example, all Qwen models without extrapolation favor $(K=4096, \beta=2.0)$, while Qwen models with YaRN consistently favor $(K=8192, \beta=2.5)$.

\subsection{Choosing Parameters for \attnts{}}
\label{app:param_attns}

We select the temperature parameter $\tau$ for \attnts{} using the same \mrcr{}-based protocol. We search over $\tau \in \{0.975, 0.95, 0.9, 0.85\}$. Due to differences in attention sharpness across model families and scales, the optimal $\tau$ varies slightly:
\[
\tau = 0.95 \text{ (Qwen3-4B)}, \;
\tau = 0.975 \text{ (Qwen3-8B)}, \; 
\tau = 0.975 \text{(Qwen3-32B)}, \;
\tau = 0.9 \text{(LLaMA)}.
\]
These values are fixed across all tasks and context lengths for each model.

\section{Robustness to Hyperparameter Variations}
\label{app:param_robustness}

\begin{table}[h]
\centering
\footnotesize
\renewcommand{\tabcolsep}{1.5mm}
\begin{tabular}{lccc|cc}
\toprule
 & \textbf{$p$} & \textbf{$K$} & $\boldsymbol{\beta}$ 
 & \textbf{MRCR} & \longbench{} \\
\midrule
Vanilla & -- & -- & -- & 16.1 & 28.6 \\

\midrule
\textbf{Default} & 0.975 & 8192 & 2.5 & 19.6 & 34.7 \\

\midrule
\multicolumn{6}{l}{\textit{Vary $p$}} \\
 & 0.95 & 8192 & 2.5 & 18.4 & 31.7 \\

\midrule
\multicolumn{6}{l}{\textit{Varying $K$}} \\
 & 0.975 & 2048 & 2.5 &\textbf{ 19.6 }& \textbf{36.8} \\
 & 0.975 & 4096 & 2.5 & 19.0 & 32.7 \\

\midrule
\multicolumn{6}{l}{\textit{Varying $\beta$}} \\
 & 0.975 & 8192 & 3.0 & 18.5 & 34.7 \\
 & 0.975 & 8192 & 3.5 & 18.1 & 31.7 \\
\bottomrule
\end{tabular}
\caption{Evaluation of hyperparameter robustness of \methodname{} with Qwen3-8B. \methodname{} is robust to modest variations in the rescale strength $\beta$, selection constraint $K$, and selection threshold $p$. }
\label{tab:hypersense}
\end{table}

We evaluate the robustness of \methodname{} to hyperparameter variations on \mrcr{}-128K and \longbench{}-128K using Qwen3-8B. We vary one hyperparameter at a time while fixing the others to the default configuration.

Table~\ref{tab:hypersense} shows that \methodname{} \textbf{consistently outperforms vanilla decoding across all tested settings}. While more aggressive configurations (e.g., $\beta=3.5$) lead to smaller gains, they still outperform the vanilla baseline.

Importantly, we observe a \textbf{strong correspondence between performance on \mrcr{} and \longbench{}}: hyperparameters that perform well on the synthetic \mrcr{} benchmark also tend to yield strong improvements on \longbench{}. This trend indicates that the effects of hyperparameter choices are consistent across tasks, and that parameters selected on a synthetic long-context benchmark generalize well to more realistic long-context reasoning tasks.

\section{Details of the Path Traversal Task }
\label{app:detail_qr_analysis}

In \S\ref{sec:qrhead_analysis}, we use \pathtrans{} to stress-test the basic long-context reasoning capability of LMs. The input to \pathtrans{} consists of a long list of graph edges $E=\{\langle v_i, v_j\rangle\}$ between nodes. The task is to find a path $\mathcal{T} = (\langle v_{\text{start}},v_1\rangle\, \ldots,\langle v_{t-1}, v_{\text{target}}\rangle)$ connecting a given start node to a target node.
Crucially, the graph is designed in a way that each node along the gold path has \emph{exactly one} outgoing edge. As a result, solving the task reduces to repeatedly identifying the next correct outgoing node and chaining them together. This constraint is \textit{explicitly stated} in the prompt, ensuring a deterministic reasoning procedure.
This design eliminates variation due to search order while preserving the core difficulty of long-context reasoning: repeatedly retrieving key information from a long and semantically dense context at multiple decoding steps.
By varying the number of nodes, we can control the input length and construct tasks with different context sizes. We provide a full prompt in Prompt~\ref{prompt:eval_path_travers}.

Specifically, we generate instances with approximately 4, 8, 16, and 32K tokens, corresponding to roughly 250, 500, 1000, and 2000 edges, respectively. LMs are required to find a path of 5 nodes (four edges).
As shown in Figure~\ref{fig:motivation} (left), although the task is structurally simple, model performance degrades rapidly as context length increases to 32K, well below the claimed context size, with step-level accuracy (correctness of each $\langle v_{t-1}, v_{t}\rangle$ along the path) dropping from near-perfect to approximately 20\%.
\pathtrans{} thus isolates a central challenge of long-context reasoning: the need for repeated dynamic key-context lookup during decoding.

\begin{prompt}[title={Prompt \thetcbcounter: Prompt for the Path Traversal Task}, label=prompt:eval_path_travers]
\prompttext{ $[$TASK$]$\\
In a completely hypothetical world, there are a number of cities. Each city has a one-way connection to only one other city via a specific transit method (bus, train, plane, or ferry). Your task is to provide a route from a city to another city. You should follow the specific instruction provided later and output the route following the format provided in the instruction.\\ \\
$[$IMPORTANT NOTES$]$\\
- All connections are one-way. If city A is connected to city B, you can travel from A to B, but not the other way around.\\
- Because each city is connected to only one other city, so there's only one possible route. To find the route, you can simply start from the starting city, identify the next city it's connected to, and repeat the process until you reach the destination city.\\
- Please follow the exact format specified below when outputting the route.\\ \\
$[$OUTPUT FORMAT$]$\\
Please mark the route with \textless Route\textgreater and \textless /Route\textgreater tags. The route should be in the following format, where one line is one step of the route:\\
\textless Route\textgreater\\
From \textless CITY\_NAME\textgreater, take a \textless TRANSIT\_METHOD\textgreater to \textless CITY\_NAME\textgreater.\\
...\\
From \textless CITY\_NAME\textgreater, take a \textless TRANSIT\_METHOD\textgreater to \textless CITY\_NAME\textgreater.\\
\textless /Route\textgreater\\ \\
$[$EXAMPLE$]$\\
In a hypothetical world, there are a number of cities. Each city has a one-way connection to only one other city via a specific transit method. The details of the cities are as follows:\\ \\
Fort Worth is a lively city. You can travel from Fort Worth to Manchester by ferry.\\
Leeds is a lively city. You can travel from Leeds to London by bus.\\
Manchester is a lively city. You can travel from Manchester to Indianapolis by plane.\\
Houston is a lively city. You can travel from Houston to London by ferry.\\
Charlotte is a lively city. You can travel from Charlotte to Charlotte by bus.\\
London is a lively city. You can travel from London to San Antonio by train.\\
San Antonio is a lively city. You can travel from San Antonio to Kitchener by train.\\
Seattle is a lively city. You can travel from Seattle to London by train.\\
Indianapolis is a lively city. You can travel from Indianapolis to Houston by ferry.\\ \\
Now find the route from Manchester to Kitchener based on the information above.\\ \\
\textless Route\textgreater \\
From Manchester, take a plane to Indianapolis.\\
From Indianapolis, take a ferry to Houston.\\
From Houston, take a ferry to London.\\
From London, take a train to San Antonio.\\
From San Antonio, take a train to Kitchener.\\
\textless /Route\textgreater\\ \\
$[$PROBLEM$]$\\
\param{problem context} \\
Now find the route from \param{src city} to \param{dst city} based on the information above. Some reminders:\\
- All connections are one-way. You can solve the problem by iteratively finding the next city to travel to until you reach the destination city.\\
- Follow the specific format for the route output. Mark the route with \textless Route\textgreater and \textless /Route\textgreater tags.\\
}
\end{prompt}

\section{Qualitative Examples}
\label{app:qualitative_example}

\definecolor{taskbg}{RGB}{235, 240, 255}
\definecolor{vanillaframe}{RGB}{195, 110, 110}
\definecolor{vanillabg}{RGB}{255, 248, 248}
\definecolor{discoframe}{RGB}{80, 155, 115}
\definecolor{discobg}{RGB}{244, 252, 248}
\definecolor{refframe}{RGB}{80, 100, 160}
\definecolor{structbg}{RGB}{248, 248, 248}

\newcommand{\wulq}{\bgroup
  \markoverwith{\textcolor{vanillaframe}{\rule[-.8ex]{.3em}{1.2pt}}}%
  \ULon}
\newcommand{\culq}{\bgroup
  \markoverwith{\textcolor{discoframe}{\rule[-.8ex]{.3em}{1.2pt}}}%
  \ULon}
\newcommand{\annbad}[1]{\,\textcolor{gray!75!black}{\scriptsize(\textcolor{red!70!black}{\ding{55}}~#1)}}
\newcommand{\annok}[1]{\,\textcolor{gray!75!black}{\scriptsize(\textcolor{green!60!black}{\ding{51}}~#1)}}
\newcommand{\roleUser}{\textbf{User:}}
\newcommand{\roleAsst}{\textbf{Assistant:}}
\newcommand{\roleAsstRef}{\textcolor{refframe}{\textbf{Assistant:}}}
\newcommand{\otherturnssep}{\hfill\textcolor{gray!70}{$\vdots$\;\footnotesize other turns}\hfill}

We present qualitative examples illustrating how \methodname{} improves long-context reasoning in LMs.

In Figure~\ref{fig:qualitative-path}, we show a \pathtrans{} example where vanilla decoding produces an incorrect output, while \methodname{} yields the correct result. In Figure~\ref{fig:herne-attention}, we visualize the averaged attention patterns at the first error step, comparing \qrhead{} with all attention heads. The results highlight the robustness of \qrhead{} in consistently attending to relevant context for retrieval.

We further present two failure cases of vanilla decoding on \mrcr{} (Figures~\ref{fig:qualitative-wrong-turn} and~\ref{fig:qualitative-imprecise}), both of which are partially corrected by \methodname{}.

\begin{figure}[t]
\scriptsize
\begin{tcolorbox}[
  colframe=gray!75, colback=white,
  colbacktitle=gray!75, coltitle=white,
  boxrule=0.6pt, arc=3pt,
  left=5pt, right=5pt, top=3pt, bottom=3pt,
  title={\small\textbf{Path Traversal: Route Hallucination}},
  fonttitle=\small\bfseries,
]

\begin{tcolorbox}[
  title={\footnotesize\textbf{Context} \textcolor{gray}{(16k tokens, graph description)}},
  colback=structbg, colframe=gray!20,
  fonttitle=\footnotesize, boxrule=0.2pt, arc=2pt,
  left=5pt, right=5pt, top=3pt, bottom=3pt,
]
Yaroslavl is a lively city. You can travel from Yaroslavl to Hayward by train.%
  \hfill\textcolor{gray!70!black}{\tiny AllHead \#1}\\
\ldots\\
Kazan is a lively city. You can travel from Kazan to Sarnia by ferry.%
  \hfill\textcolor{gray!70!black}{\tiny AllHead \#2}\\
\ldots\\
Newmarket is a lively city. You can travel from Newmarket to Thornton by ferry.%
  \hfill\textcolor{discoframe}{\tiny QRHead \#2}\\
\textcolor{refframe}{\textbf{Thornton is a lively city. You can travel from Thornton to Leatherhead by ferry.}}%
  \textcolor{refframe}{\scriptsize~$\checkmark$~\textbf{needle at the first error}}\hfill\textcolor{discoframe}{\tiny QRHead \#1}\\
Thornton is a lively city. You can travel from Thornton to Bordon by ferry.\\
Bordon is a lively city. You can travel from Bordon to Warsaw by ferry.%
  \hfill\textcolor{discoframe}{\tiny QRHead \#3}\\
\ldots\\
Halesowen is a lively city. You can travel from Halesowen to Spokane by bus.%
  \hfill\textcolor{gray!70!black}{\tiny AllHead \#3}\\
\ldots\\[2pt]
\colorbox{taskbg}{\parbox{\dimexpr\linewidth-2\fboxsep\relax}{%
  \roleUser{} Find the route from \textbf{Herne} to \textbf{Leatherhead}.%
  \hfill\textcolor{gray}{\scriptsize(task)}%
}}
\end{tcolorbox}

\vspace{3pt}
\begin{tcolorbox}[
  title={\footnotesize\textcolor{vanillaframe}{\textbf{Vanilla}}
         \textcolor{gray}{(accuracy: 0.0)}},
  colframe=vanillaframe!50, colback=vanillabg,
  colbacktitle=vanillaframe!15, coltitle=black,
  fonttitle=\footnotesize, boxrule=0.3pt, arc=2pt,
  left=4pt, right=4pt, top=2pt, bottom=2pt,
]
From Herne, take a plane to Garden Grove.\\
From Garden Grove, take a bus to Newmarket.\\
From Newmarket, take a ferry to Thornton.\\
\wulq{From Thornton, take a ferry to Bordon.}\annbad{wrong destination}%
  \hfill\textcolor{vanillaframe}{\scriptsize$\leftarrow$~error starts}\\
From Bordon, take a ferry to Gloucester.\\
From Gloucester, take a train to Reno.\\
\ldots\ (never reaches Leatherhead)\annbad{path drift}
\end{tcolorbox}

\vspace{3pt}
\begin{tcolorbox}[
  title={\footnotesize\textcolor{discoframe}{\textbf{DySCO (ours)}}
         \textcolor{gray}{(accuracy: 1.0)}},
  colframe=discoframe!50, colback=discobg,
  colbacktitle=discoframe!15, coltitle=black,
  fonttitle=\footnotesize, boxrule=0.3pt, arc=2pt,
  left=4pt, right=4pt, top=2pt, bottom=2pt,
]
From Herne, take a plane to Garden Grove.\\
From Garden Grove, take a bus to Newmarket.\\
From Newmarket, take a ferry to Thornton.\\
\culq{From Thornton, take a ferry to Leatherhead.}\annok{correct destination}
\end{tcolorbox}

\end{tcolorbox}
\caption{%
  Path Traversal example (Herne $\to$ Leatherhead, 16k context).
  The context box shows the top-3 attended lines at the first error step
  under QRHeads (green) and all heads (gray): QRHeads rank the needle edge
  (Thornton $\to$ Leatherhead) first, while all heads attend to
  irrelevant cities. Both vanilla and DySCO decoding follow the first three
  steps correctly; at step~4, vanilla decoding takes the wrong edge
  (Thornton $\to$ Bordon) and drifts indefinitely (accuracy 0.0), whereas
  \methodname{}---guided by QRHead attention---retrieves the correct edge
  and terminates on target (accuracy 1.0).%
}
\label{fig:qualitative-path}
\end{figure}

\begin{figure}[t]
  \centering
  \includegraphics[width=\linewidth]{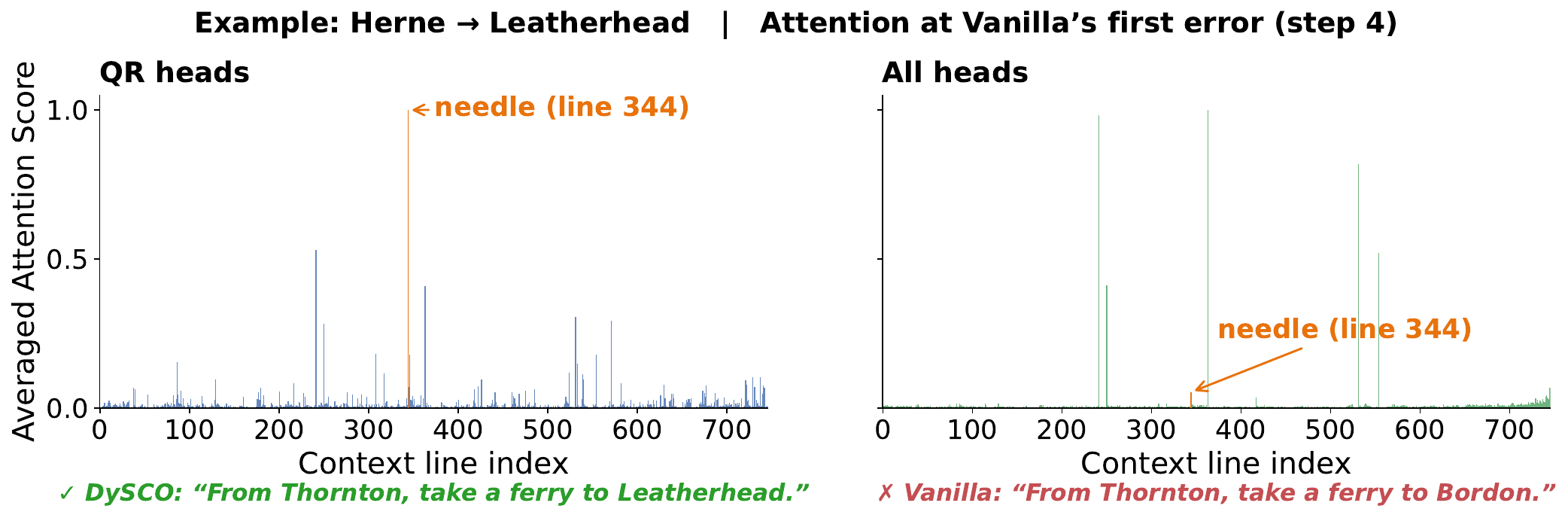}
  \caption{%
    Attention distribution at the first error step (step 4) of the
    Herne $\to$ Leatherhead \pathtrans{} example, where the model must
    generate ``From Thornton, take a ferry to Leatherhead.''
    \textbf{Left (QRHeads):} attention is highly concentrated on the needle
    line describing the Thornton $\to$ Leatherhead connection, which ranks
    first; \methodname{} leverages this signal and generates the correct step.
    \textbf{Right (all heads):} the average attention across all heads is
    more dispersed and the needle line's rank drops significantly, causing
    vanilla decoding to misidentify the next city and generate
    ``From Thornton, take a ferry to Bordon'' instead.
    This illustrates that \qrhead{}s inherently possess stronger retrieval
    signal focus \emph{without} any rescaling intervention, and \methodname{}
    amplifies these signals to guide generation correctly where vanilla fails.%
  }
  \label{fig:herne-attention}
\end{figure}

\begin{figure}[t]
\scriptsize
\renewcommand{\otherturnssep}{\hfill\textcolor{gray!70}{$\vdots$\;\scriptsize other turns}\hfill}
\begin{tcolorbox}[
  colframe=gray!75, colback=white,
  colbacktitle=gray!75, coltitle=white,
  boxrule=0.6pt, arc=3pt,
  left=6pt, right=6pt, top=4pt, bottom=4pt,
  title={\small\textbf{MRCR Failure Mode 1: Wrong Turn Retrieval}},
  fonttitle=\small\bfseries,
]

\begin{tcolorbox}[
  title={\footnotesize\textbf{Context} \textcolor{gray}{(242 turns)}},
  colback=structbg, colframe=gray!20,
  fonttitle=\footnotesize, boxrule=0.2pt, arc=2pt,
  left=5pt, right=5pt, top=3pt, bottom=3pt,
]
\roleUser{} Write a short essay about individuals.\\
\roleAsst{} Individual Responsibility in Modern Society\ldots
  \hfill\textcolor{gray}{\scriptsize(1st essay)}\\[2pt]
\otherturnssep\\[2pt]
\roleUser{} Write a short essay about individuals.\\
\roleAsst{} The Strength of the Individual\ldots
  \hfill\textcolor{gray}{\scriptsize(2nd essay)}\\[2pt]
\otherturnssep\\[2pt]
\roleUser{} Write a short essay about individuals.\\
\roleAsst{} The Power and Complexity of Individuals\ldots
  \hfill\textcolor{gray}{\scriptsize(3rd essay)}\;\textcolor{vanillaframe}{\scriptsize$\leftarrow$~vanilla retrieves this}\\[2pt]
\otherturnssep\\[2pt]
\roleUser{} Write a short essay about individuals.\\
\roleAsstRef{} \textbf{Title: The Unique Impact of Individuals on Society}
  \hfill\textcolor{refframe}{$\checkmark$~\textbf{reference (4th essay)}}\\
\phantom{\roleAsstRef{}~}In the tapestry of human civilization, individuals act as
  the distinct threads\ldots\\[2pt]
\otherturnssep\\[2pt]
\colorbox{taskbg}{\parbox{\dimexpr\linewidth-2\fboxsep\relax}{%
  \roleUser{} Prepend \texttt{[X]} to the \textbf{4th} essay about \textbf{individuals}.%
  \hfill\textcolor{gray}{\scriptsize(task)}%
}}
\end{tcolorbox}

\vspace{4pt}
\begin{tcolorbox}[
  title={\footnotesize\textcolor{vanillaframe}{\textbf{Vanilla}}
         \textcolor{gray}{(match ratio: 0.048)}},
  colframe=vanillaframe!50, colback=vanillabg,
  colbacktitle=vanillaframe!15, coltitle=black,
  fonttitle=\footnotesize, boxrule=0.3pt, arc=2pt,
  left=4pt, right=4pt, top=2pt, bottom=2pt,
]
\begin{tcolorbox}[
  colback=gray!7, colframe=gray!30,
  boxrule=0.3pt, arc=1pt,
  left=3pt, right=3pt, top=1pt, bottom=1pt,
  title={\scriptsize\texttt{<think>}}, fonttitle=\scriptsize,
]
\scriptsize\itshape
\ldots the fourth essay was titled
\wulq{``The Power and Complexity of Individuals''}\annbad{wrong turn identified}
\ldots
\end{tcolorbox}
\vspace{3pt}
\wulq{The Power and Complexity of Individuals}\annbad{wrong essay title}\\[2pt]
In the tapestry of human \wulq{society}\annbad{civilization},
individuals are the \wulq{threads that weave together the intricate patterns}
of communities, cultures, and nations.
Each person, unique in their own right,
contributes distinct perspectives, experiences, and talents
that build the dynamic structure of our world\ldots
\end{tcolorbox}

\vspace{3pt}
\begin{tcolorbox}[
  title={\footnotesize\textcolor{discoframe}{\textbf{DySCO (ours)}}
         \textcolor{gray}{(match ratio: 1.000)}},
  colframe=discoframe!50, colback=discobg,
  colbacktitle=discoframe!15, coltitle=black,
  fonttitle=\footnotesize, boxrule=0.3pt, arc=2pt,
  left=4pt, right=4pt, top=2pt, bottom=2pt,
]
\begin{tcolorbox}[
  colback=gray!7, colframe=gray!30,
  boxrule=0.3pt, arc=1pt,
  left=3pt, right=3pt, top=1pt, bottom=1pt,
  title={\scriptsize\texttt{<think>}}, fonttitle=\scriptsize,
]
\scriptsize\itshape
\ldots The fourth essay\ldots was about
\culq{the unique impact of individuals on society}\annok{correct turn identified}
\ldots
\end{tcolorbox}
\vspace{3pt}
\culq{Title: The Unique Impact of Individuals on Society}\annok{correct essay retrieved}\\[2pt]
In the tapestry of human \culq{civilization},
individuals act as the \culq{distinct threads} that, when woven together,
create the intricate and \culq{colorful fabric} of society.
These unique beings, each with their own experiences, perspectives, and abilities,
contribute to the collective advancement and evolution of the human race\ldots
\end{tcolorbox}

\end{tcolorbox}
\caption{%
  Failure Mode 1 on \mrcr{}: wrong turn retrieval (64k context, 242 turns).
  The context embeds four essays about ``individuals''; vanilla decoding mistakes
  the 3rd essay for the 4th (match ratio 0.048), while \methodname{} correctly
  identifies and retrieves the target turn (match ratio 1.000).%
}
\label{fig:qualitative-wrong-turn}
\end{figure}

\begin{figure}[t]
\scriptsize
\renewcommand{\otherturnssep}{\hfill\textcolor{gray!70}{$\vdots$\;\scriptsize other turns}\hfill}
\begin{tcolorbox}[
  colframe=gray!75, colback=white,
  colbacktitle=gray!75, coltitle=white,
  boxrule=0.6pt, arc=3pt,
  left=3pt, right=3pt, top=1pt, bottom=1pt,
  title={\small\textbf{MRCR Failure Mode 2: Imprecise Copying}},
  fonttitle=\small\bfseries,
]

\begin{tcolorbox}[
  title={\footnotesize\textbf{Context} \textcolor{gray}{(156 turns)}},
  colback=structbg, colframe=gray!20,
  fonttitle=\footnotesize, boxrule=0.2pt, arc=2pt,
  left=5pt, right=5pt, top=3pt, bottom=3pt,
]
\roleUser{} Write a short scene in a play about application.\\
\roleAsst{} The Negotiation\ldots
  \hfill\textcolor{gray}{\scriptsize(1st scene)}\\
\otherturnssep\\
\roleUser{} Write a short scene in a play about application.\\
\roleAsst{} The Presentation\ldots
  \hfill\textcolor{gray}{\scriptsize(2nd scene)}\\
\otherturnssep\\
\roleUser{} Write a short scene in a play about application.\\
\roleAsst{} The Follow-Up\ldots
  \hfill\textcolor{gray}{\scriptsize(3rd scene)}\\
\otherturnssep\\
\roleUser{} Write a short scene in a play about application.\\
\roleAsstRef{} \textbf{Title: The Interview}
  \hfill\textcolor{refframe}{$\checkmark$~\textbf{reference (4th scene)}}\\
\phantom{\roleAsstRef{}~}\ldots\\
\phantom{\roleAsstRef{}~}\textbf{EMMA}: It would have to be seeing my ideas come to life\ldots\ diverse perspectives.\\
\phantom{\roleAsstRef{}~}\textbf{MR.\ JENKINS}: That's one of the crucial things---teamwork\ldots\\
\phantom{\roleAsstRef{}~}\textbf{EMMA}: Certainly! I'm proficient in Google Analytics\ldots\\
\phantom{\roleAsstRef{}~}\textbf{MR.\ JENKINS}: What's a challenge you've faced?\\
\phantom{\roleAsstRef{}~}\textbf{EMMA}: During a campaign, our engagement metrics\ldots\\
\phantom{\roleAsstRef{}~}\textbf{MR.\ JENKINS}: Problem-solving and adaptability\ldots\ Before concluding, is there anything you'd like to ask?\\
\phantom{\roleAsstRef{}~}\textbf{EMMA}: What's the most rewarding part about working here?\\
\phantom{\roleAsstRef{}~}\ldots\\
\otherturnssep\\
\colorbox{taskbg}{\parbox{\dimexpr\linewidth-2\fboxsep\relax}{%
  \roleUser{} Prepend \texttt{[X]} to the \textbf{4th} scene in a play about \textbf{application}.%
  \hfill\textcolor{gray}{\scriptsize(task)}%
}}
\end{tcolorbox}

\vspace{2pt}
\begin{tcolorbox}[
  title={\footnotesize\textcolor{vanillaframe}{\textbf{Vanilla}}
         \textcolor{gray}{(match ratio: 0.542)}},
  colframe=vanillaframe!50, colback=vanillabg,
  colbacktitle=vanillaframe!15, coltitle=black,
  fonttitle=\footnotesize, boxrule=0.3pt, arc=2pt,
  left=4pt, right=4pt, top=2pt, bottom=2pt,
]
\begin{tcolorbox}[
  colback=gray!7, colframe=gray!30,
  boxrule=0.3pt, arc=1pt,
  left=3pt, right=3pt, top=1pt, bottom=1pt,
  title={\scriptsize\texttt{<think>}}, fonttitle=\scriptsize,
]
\scriptsize\itshape
\ldots The 4th scene\ldots was titled
\culq{``The Interview''}\annok{correct turn identified}
with characters MR.\ JENKINS and EMMA\ldots
\end{tcolorbox}
\vspace{0pt}
Title: The Interview\\
\ldots\\
\textbf{EMMA}: It would have to be seeing my ideas come to life\ldots\ diverse perspectives.\\
\textbf{MR.\ JENKINS}: Problem-solving\ldots\ What's a challenge?\annbad{wrong order}%
  \hfill\textcolor{vanillaframe}{\scriptsize$\leftarrow$~error starts}\\
\textbf{EMMA}: During a campaign, our engagement metrics\ldots\annbad{wrong order}\\
\textbf{MR.\ JENKINS}: That's one of the crucial things---teamwork\ldots\\
\textbf{EMMA}: Certainly! I'm proficient in Google Analytics\ldots\\
\textbf{MR.\ JENKINS}: What's a challenge?\annbad{duplicated}\\
\textbf{EMMA}: During a campaign\ldots\annbad{duplicated}\\
\textbf{MR.\ JENKINS}: teamwork\ldots\annbad{duplicated}\\
\textbf{EMMA}: Certainly! Google Analytics\ldots\annbad{duplicated}\\
\textbf{MR.\ JENKINS}: Excellent\ldots\ Before concluding, is there anything you'd like to ask?\\
\textbf{EMMA}: Yes\ldots\ What's the most rewarding part about working here?\\
\ldots
\end{tcolorbox}

\vspace{0pt}
\begin{tcolorbox}[
  title={\footnotesize\textcolor{discoframe}{\textbf{DySCO (ours)}}
         \textcolor{gray}{(match ratio: 0.750)}},
  colframe=discoframe!50, colback=discobg,
  colbacktitle=discoframe!15, coltitle=black,
  fonttitle=\footnotesize, boxrule=0.3pt, arc=2pt,
  left=4pt, right=4pt, top=2pt, bottom=2pt,
]
\begin{tcolorbox}[
  colback=gray!7, colframe=gray!30,
  boxrule=0.3pt, arc=1pt,
  left=3pt, right=3pt, top=1pt, bottom=1pt,
  title={\scriptsize\texttt{<think>}}, fonttitle=\scriptsize,
]
\scriptsize\itshape
\ldots there was a scene titled
\culq{``The Interview''}\annok{correct turn identified}
where Emily applies for a job\ldots
\end{tcolorbox}
\vspace{0pt}
Title: The Interview\\
\ldots\\
\textbf{EMMA}: It would have to be seeing my ideas come to life\ldots\ diverse perspectives.\\
\textbf{MR.\ JENKINS}: Problem-solving\ldots\ What's a challenge?\annbad{wrong order}\\
\textbf{EMMA}: During a campaign, our engagement metrics\ldots\annbad{wrong order}\\
\textbf{MR.\ JENKINS}: That's one of the crucial things---teamwork\ldots\\
\textbf{EMMA}: Certainly! I'm proficient in Google Analytics\ldots\\
\textbf{MR.\ JENKINS}: Excellent\ldots\ Before concluding, is there anything you'd like to ask?\\
\textbf{EMMA}: Yes\ldots\ What's the most rewarding part about working here?\ \textcolor{gray}{(\textcolor{green!60!black}{\ding{51}}~no duplication)}\\
\ldots
\end{tcolorbox}

\end{tcolorbox}
\caption{%
  Failure Mode 2 on \mrcr{}: imprecise copying (64k context, 156 turns).
  Both vanilla and \methodname{} retrieve the correct turn (Title: The Interview)
  but misorder the dialogue.
  Vanilla further corrupts the output by duplicating a full block of dialogue
  (match ratio 0.542). \methodname{} avoids the repetition, reproducing the
  content more faithfully (match ratio 0.750).%
}
\label{fig:qualitative-imprecise}
\end{figure}

\end{document}